\documentclass[acmtog,authorversion]{acmart}

\AtBeginDocument{%
  \providecommand\BibTeX{{%
    \normalfont B\kern-0.5em{\scshape i\kern-0.25em b}\kern-0.8em\TeX}}}

\setcopyright{acmcopyright}\acmJournal{TOG}
\acmYear{2022}\acmVolume{41}\acmNumber{6}\acmArticle{186}\acmMonth{12} \acmDOI{10.1145/3550454.3555440}

\usepackage{enumerate}
\usepackage{multirow} 

\begin{document}

\title[SHRED: 3D Shape Region Decomposition with Learned Local Operations]{SHRED: 3D Shape Region Decomposition with \\ Learned Local Operations}

\author{R. Kenny Jones}
\email{russell\_jones@brown.edu}
\affiliation{%
    \institution{Brown University}
    \country{USA}
}

\author{Aalia Habib}
\email{aalia\_habib@brown.edu}
\affiliation{%
    \institution{Brown University}
    \country{USA}
}

\author{Daniel Ritchie}
\email{daniel\_ritchie@brown.edu}
\affiliation{%
    \institution{Brown University}
    \country{USA}
}

\begin{abstract}

We present SHRED, a method for 3D SHape REgion Decomposition. 
SHRED takes a 3D point cloud as input and uses learned local operations to produce a segmentation that approximates fine-grained part instances. 
We endow SHRED with three decomposition operations: splitting regions, fixing the boundaries between regions, and merging regions together.
Modules are trained independently and locally, allowing SHRED to generate high-quality segmentations for categories not seen during training.
We train and evaluate SHRED with fine-grained segmentations from PartNet; using its merge-threshold hyperparameter, we show that SHRED produces segmentations that better respect ground-truth annotations compared with baseline methods, at any desired decomposition granularity. 
Finally, we demonstrate that SHRED is useful for downstream applications, out-performing all baselines on zero-shot fine-grained part instance segmentation and few-shot fine-grained semantic segmentation when combined with methods that learn to label shape regions.

\end{abstract}



\begin{teaserfigure}
  \includegraphics[width=\textwidth]{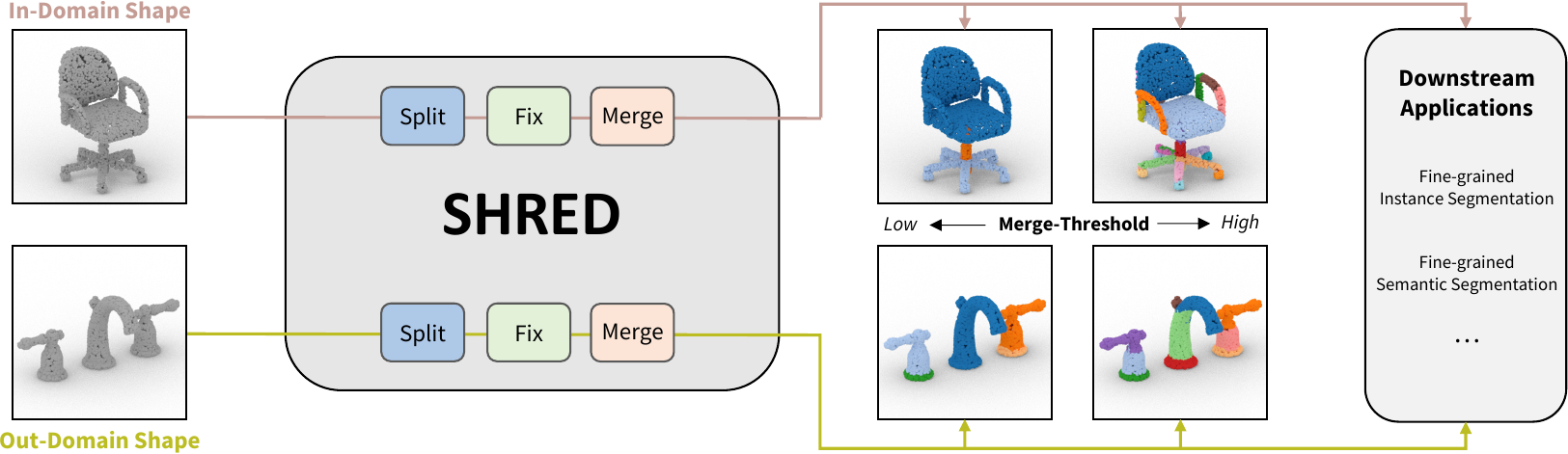}
  \caption{SHRED decomposes 3D shapes into regions by learning to locally split, fix, and merge segments. A merge-threshold parameter can be adjusted to change decomposition granularity depending on the target downstream application.}
  \label{fig:teaser}
\end{teaserfigure}

\maketitle

\section{Introduction}

3D segmentation is a fundamental problem within computer graphics and vision.
Many applications in these fields require methods capable of decomposing 3D shapes and scenes into meaningful regions:
establishing correspondences,
skeleton extraction,
guiding semantic labeling,
and allowing autonomous agents to interact with objects, 
to name a few \cite{specClustSeg, landrieu2018large, BenKinematics}.

While there is a long history of computer graphics research using geometric and visual cues to segment 3D shapes into regions, most recent work has explored the use of data-driven neural methods.  
Learning-based approaches have produced impressive results on coarse decomposition tasks where training annotations are plentiful, but often struggle when either of these conditions is not met.
Most methods that operate within this paradigm learn to segment shapes globally, which contributes to both of these limitations.
While coarse decompositions are often fairly globally consistent, fine-grained decomposition are highly varied, even within the same category (e.g. chairs almost always have a back, a seat, and a base, but only a small subset of chairs have wheel casters). 
Moreover, these methods are often unable to produce sensible segmentations for shapes outside of their training distribution, as their networks specialize to the global patterns of training shapes.

Some recent approaches have aimed to develop learning-based systems capable of producing fine-grained segmentations in a way that generalizes across domains.
Within this paradigm, the hope is to train networks on categories of shapes with abundant annotations, and then show these networks have learned decomposition policies that can be directly applied to shapes from categories that lack annotations.
Many of these methods use at least one network that reasons locally, with the idea that forcing the network to ignore global context might encourage better generalization on out-domain shapes \cite{Wang_2021_CVPR, Luo2020Learning}.

Following this general framework, SHRED learns a collection of region decomposition operations that all reason in a local fashion: splitting a region into sub-regions, fixing the boundaries between regions, and deciding when neighboring regions should be merged together.
Each of these operations is modeled with a neural module that operates over regions, represented as point-clouds, and trained with ground-truth part annotations from a dataset of shapes.
The modules of SHRED are applied sequentially to generate a region decomposition.
An input shape first undergoes a naive decomposition using farthest-point sampling, then these regions are passed through the split, fix and merge operators to produce the final segmentation.
Of note, the merge operator exposes a merge-threshold hyperparameter that can be toggled to control the granularity of the output decomposition.

We train a version of SHRED on three data abundant categories of PartNet: chairs, lamps and storage furniture (\textit{in-domain} shapes), and evaluate its ability to produce region decompositions for test-set in-domain and out-domain shapes.
We compare SHRED against other shape segmentation methods, including learned and non-learned approaches, that operate both globally and locally.
We find that SHRED produces better region decompositions than baseline methods, for both in-domain and out-domain categories.
Analyzing the trade-off between decomposition quality and granularity, we vary SHRED's merge-threshold to create a Pareto frontier of solutions that strictly dominates all comparisons.
We then demonstrate that SHRED's decompositions can be treated as fine-grained part instance segmentations that outperform comparison methods.
Finally, we evaluate how SHRED can improve fine-grained semantic segmentation, using the output of SHRED as the input to a method that learns to semantically label shape regions, and find that using SHRED results in the best performance.

In summary, our contributions are:
\begin{enumerate}[(i)]
    \item SHRED, a method for 3D SHape REgion Decomposition with learned local split, fix and merge operations.  
    \item Demonstrations on collections of manufactured shapes that 
    SHRED outperforms baseline methods 
    in terms of
    fine-grained segmentation performance and
    finding better trade-offs between decomposition quality and granularity, 
    for both in-domain and out-domain categories.
\end{enumerate}

Code for our method and experiments can be found at found at https://github.com/rkjones4/SHRED .

\section{Related Work}

\definecolor{lblue}{RGB}{219, 227, 241}
\definecolor{lgreen}{RGB}{229, 240, 220}
\definecolor{lorange}{RGB}{248, 230, 217}

\begin{figure*}[t!]
  \includegraphics[width=\linewidth]{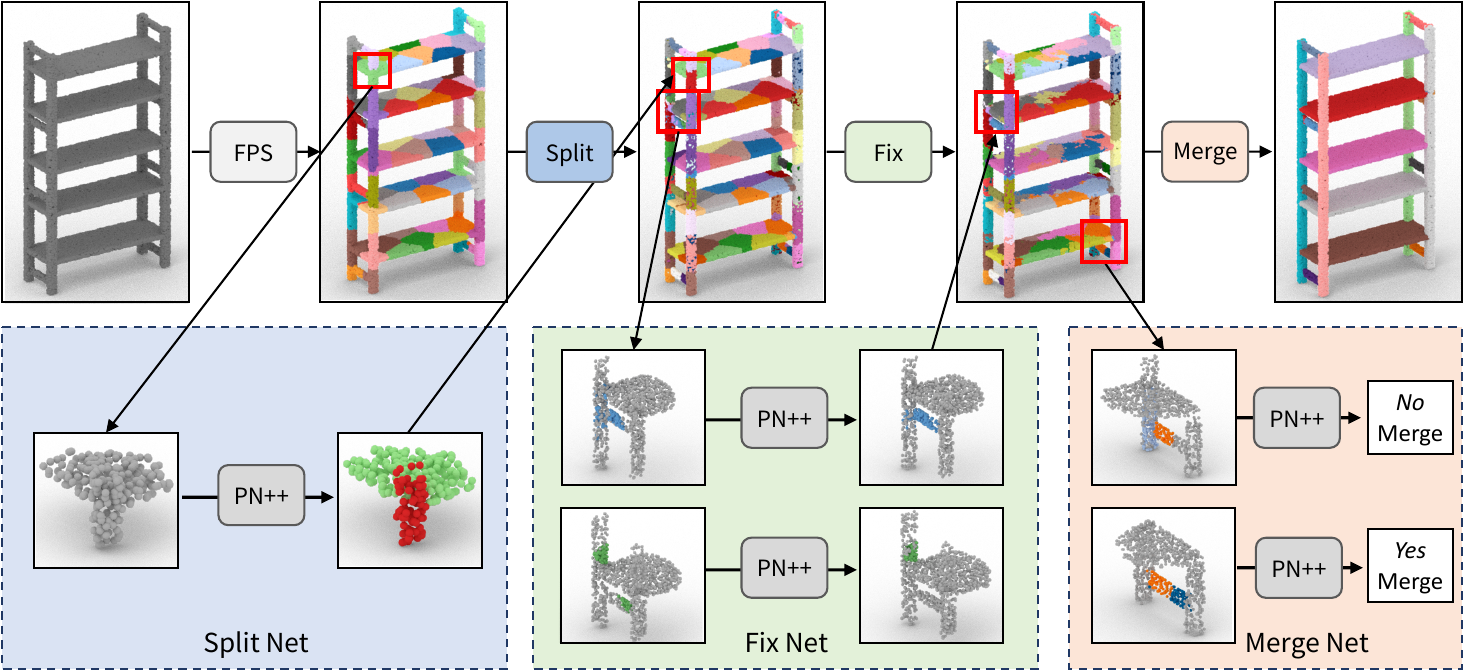}
\caption{The modules of SHRED. From left to right, input shapes are naively decomposed by farthest-point sampling (FPS), regions are \colorbox{lblue}{split} into sub-regions, boundaries are \colorbox{lgreen}{fixed}, and neighbors are \colorbox{lorange}{merged} together. Bottom-row cut-outs visualize network input-outputs. }
\label{fig:overview}
\end{figure*}

\paragraph{Shape segmentation with geometric cues} 
There is a long history of computer graphics research dedicated to segmenting a 3D shape into meaningful regions according to geometric properties.
Most such methods are not data-driven, but rather analyze heuristic properties of 3D meshes to produce shape decompositions useful for various downstream applications. 
These include approaches such as:
normalized cuts, 
symmetry cues ~\cite{wang2011symmetry},
fuzzy clustering ~\cite{katz2003hierarchical},
spectral methods \cite{asafi2013weak, specClustSeg},
and approximate convex decomposition ~\cite{lien2008approximate, kaick2014shape}. 
Please refer to ~\cite{shamir2008survey} for a survey on the topic.

\paragraph{Learning to approximate shapes with primitives} 
A great body of recent research has been devoted to learning methods that aim to coarsely approximate 3D shapes with a union of primitive structures. 
Methods differ by the type of primitive used, for instance cuboids~\cite{tulsiani2017learning,AdaptiveHierarchicalCuboidAbstraction,yang2021unsupcsa},
superquadrics~\cite{SuperquadricsRevisited}, 
convex solids~\cite{deng2020cvxnet,BSPNet},
and more general local neural functions ~\cite{NeuralParts, NeuralStarDomain, genova2019learning, chen2019bae_net}.
These methods can train on 3D shapes that lack annotations and produce segmentations with paired correspondences across different shape instances.
Without annotations, these approaches rely on global reconstruction-based losses, resulting in decompositions that well-represent coarse structures, but often ignore fine-grained regions of interest.
Relatedly, some methods use primitive decompositions to formulate self-supervised losses that augment training to improve few-shot semantic labeling \cite{selfsupacd, surfit}.

\paragraph{Shape decomposition with supervised learning}
Most learning-based shape decomposition methods operate within a supervised learning paradigm by consuming a training dataset that contains annotated regions.
Many of these works learn to globally decompose a shape in a category specific fashion for a particular type of region-annotation (e.g. part instances) \cite{qi2017pointnet,qi2017pointnet++,dgcnn,yi2019gspn,wang2018sgpn,yu2019partnet}.
These approaches achieve state-of-the-art performance when the desired decomposition is coarse and training data is plentiful for the category of interest, but do not perform as well with fine-grained segmentations or out-of-domain inputs.

Recent work has tried to address these issues by developing learning-based solutions that work locally. 
For instance, \cite{Han20GenStruct} propose a pipeline for single-image shape reconstruction via cuboid proxies that can generalize to out-of-domain shapes.
Some methods have been designed with fine-grained regions in mind.
\cite{Wang_2021_CVPR} use a learned clustering approach to perform local split operations, and then aggregate sub-regions with a global merge step. 
\cite{Luo2020Learning} use policy gradient reinforcement learning to train a network that takes an over-segmented 3D shape and merges segments together to approximate a  part instance segmentation.
SHRED shares similar motivations to these last two methods: learn local operations on shape categories with abundant annotations, and then show this promotes strong generalization capabilities on novel shape categories.
We compare SHRED against these approaches, and find that SHRED's collection of learned local operations improves the quality of its region decompositions.

Related to this problem of zero-shot generalization, some methods frame shape region decomposition as a co-segmentation problem, such as AdaCoSeg \cite{AdaCoSeg}.
In this paradigm, a single labeled exemplar is provided as template to indicate how the rest of the distribution should be decomposed. 
The fix operator of SHRED takes inspiration from the part prior network in AdaCoSeg.

\paragraph{Learning to semantically label}
Some applications desire region decompositions where part instances are grouped by semantic properties. 
Learning-based methods that perform global category specific segmentations can be used for this task, although they struggle with fine-grained semantic decompositions or paradigms when labeled data is limited.
Recent approaches have investigated how to address this latter problem for coarse semantic labels.
When the number of shapes that contain any labels at all is limited, methods have investigated how to improve few-shot segmentation performance by incorporating self-supervised training objectives over large amounts of unlabeled data \cite{chen2019bae_net, selfsupacd, DBLP:conf/3dim/SharmaKM19, sun2022semi, PointContrast2020, Wang_AAAI_2021}, learning to morph shapes into matched templates from a small labeled collection \cite{Wang_2020_CVPR}, and framing few-shot segmentation under a meta-leaning paradigm \cite{huang_meta_3dv, yi_meta_arxiv}.
Related methods have focused on improving segmentation performance when the number of labels provided for each shape is limited  \cite{liu2021one, XuLee_CVPR20}.

Another line of investigation has looked into approaches that learn to assign semantic labels to regions of a 3D shape \cite{NGSP,HierarchicalSegLabelOnline}. 
When a good region decomposition is provided, these methods have been shown to outperform global semantic segmentation approaches, especially for fine-grained semantic labels and when labeled data is limited. 
We will show that the region decompositions created by SHRED can be combined with these approaches to improve few-shot fine-grained semantic segmentation performance.

\section{Method}
\label{sec:method}

SHRED takes a 3D shape as input and outputs a fine-grained region decomposition.
We define a region decomposition for shape $S$ as a set of regions  $R = \{r_0,r_1,...,r_n\}$ s.t $S = \bigcup_{i \in N}{r_i}$.
Given a ground-truth region decomposition $R^{*}$ (e.g. fine-grained part instance annotations produced by a human), we desire two properties of $R$.
First, each $r_i \in R$ should be a subset of some $r^{*}_{j} \in R^{*}$: $r_i \subseteq r^{*}_{j}$. 
Second, the number of regions in $R$ should not exceed the number of regions in $R^{*}$: $|R| \leq |R^{*}|$.
The first property states a desire that regions in $R$ do not cross the part boundaries defined by $R^{*}$, while the second property states that the granularity of $R$'s decomposition should not exceed the granularity of $R^{*}$'s decomposition.
Notice that these properties work against each other; decreasing $\|R\|$ increases the probability that some $r_{i} \in R$ is no longer a subset of some region of $R^{*}$, while ensuring that $r_i \subseteq r^{*}_{j} ~\forall~ i \in N $ requires that $|R| \ge |R^{*}|$.
These two properties are only simultaneously met when $R = R^{*}$, but we can use them to evaluate the goodness of any given $R$.

We provide a visual overview of our method in Figure \ref{fig:overview}.
SHRED contains three learned modules that perform region modifications: a split module that splits a region (Section \ref{sec:met_split}, blue box) , a fix module that fixes part boundaries between regions (Section \ref{sec:met_fix}, green box), and a merge module that decides when neighboring regions should be merged together (Section \ref{sec:met_merge}, orange box). 
As we are interested in using SHRED to generate decompositions for out-of-distribution shapes, we design each module to operate on only region-local information.
This encourages the operations to learn region-specific principals that generalize better than category-specific patterns \cite{Luo2020Learning}.
Following \cite{Wang_2021_CVPR}, we represent 3D shapes as high-resolution point-clouds (100k points) sampled from a mesh-surface, in order to capture fine-grained geometry.
SHRED first applies a naive strategy to decompose the shape into noisy regions, and then sequentially applies the split, fix, and merge modules to produce a high-quality part-respecting region decomposition.

In all experiments, we create the naive region decomposition with a simple farthest-point sampling (FPS) procedure with $K$ centroids (default $K = 64$). 
The split, fix and merge networks are all variants of PointNet++'s written in PyTorch \cite{qi2017pointnet++, pytorchpointnet++, paszke2017automatic}.
Further network training details and hyper-parameters are provided in the supplemental material.

\subsection{Split Module}
\label{sec:met_split}

The split module is tasked with deciding if a region should be further decomposed into multiple sub-regions. 
In SHRED, the split operation is applied over the naive region decompositions produced by FPS clustering, as each FPS-produced region might be an under-segmentation with respect to $R*$. 
We model the split operation with an instance segmentation network that considers each region individually; it consumes a point cloud representing a region and predicts an instance label for each point of the region.

\paragraph{Model Details} The split network uses a PointNet++ instance segmentation back-bone, with a per-point MLP head that predicts into 10 maximum part slots. 
The back-bone uses 4 set abstraction layers with 1024, 256, 64, 16 grouping points, 0.1, 0.2, 0.4, 0.8 radius size and 64, 128, 256, 512 feature size respectively. 
A series of 4 per-point feature propagation modules converts each 512 feature to size 128. 
Batch normalization and ReLU activations are used throughout.
A MLP head for per-point predictions uses a hidden layer with dimension of 64 to produce a prediction into 10 slots. The MLP uses ReLU activations and a 0.1 dropout.

The input region point clouds have 6 features (xyz positions and normals), are sub-sampled to 512 points, and are normalized to the unit-sphere.
The split network is trained with cross-entropy loss, which requires finding an alignment between predicted instance slots and the target instance slots. We use a variant of the Hungarian matching algorithm of Mo et al.~\shortcite{PartNet} to dynamically find the best alignment during training.
As we want the split network to remove all under-segmentation, our matching algorithm greedily encourages over-segmentations that better respect part boundaries, at the cost of using more prediction slots.  
We further describe this matching in the supplemental material.

\paragraph{Data Preparation} Training examples for the split network are sourced from part instance labeled training shapes. For each shape, we produce a naive region decomposition using FPS clustering. Then, each region in the naive decomposition contributes one input point cloud for training, where ground-truth target instances are supplied by the fine-grained annotated labels.

\subsection{Fix Module}
\label{sec:met_fix}

The fix module is responsible for improving region boundaries.
It consumes the region decomposition output by the split module. 
This decomposition is fine-grained, but might contain errors on part boundaries, as the split network has no information about the shape outside of each region.
While the fix module also operates over regions, unlike the split module, it also receives nearby points from outside the region to help to contextualize the region within the shape.
The input to the fix network is formed by a concatenation of two points clouds, one coming from the region of interest, one coming from outside the region of interest (colored versus grey points in Figure \ref{fig:overview}, bottom-middle).
The fix network then makes a binary prediction for each input point, deciding whether or not it should be inside or outside the region of interest.
SHRED applies this inside-outside prediction to the local neighborhood of points around each region, and these per-region decisions are propagated into a global region decomposition through an argmax operation.  

\paragraph{Model Details} The fix network uses a PointNet++ instance segmentation back-bone, with a per-point MLP that predicts a binary logit.
The back-bone is the same as the one for the split network, except that batch norm is not used. A MLP head for per-point predictions uses 2 hidden layers with dimensions of 64 and 32 to produce a binary prediction logit. The MLP uses ReLU activations and a 0.1 dropout.

Input point clouds are made up of 2048 inside points and 2048 outside points, where each point has 7 features (xyz positions, normals, inside-outside flag), and are normalized to the unit-sphere.
The network is trained with Binary Cross Entropy loss, where 1.0 (0.0) indicates the point is inside (outside) the region of interest.

\paragraph{Data Preparation} We employ a synthetic perturbation process to generate training data for the fix network. 
To generate a training example, we first sample a random ground-truth (GT) part instance from a random training set shape.
We then corrupt this region by randomly adding outside points into the region or removing points from the region.
Inside points from this corrupted region and nearby outside points (within 0.1 radius) are then combined to create an input point cloud.
Finally, we take the corrupted region, find the GT region it has the highest overlap with, and use the inside-outside values of this best-matching GT region to produce the target labels for the sampled points of the training example.

\subsection{Merge Module}
\label{sec:met_merge}

The merge module decides when neighboring regions should be combined together.
It consumes the output of the fix module, so typically the region decomposition contains little under-segmentation but has $|R| >> |R^{*}|$. 
Within our problem formulation, two regions should be merged if the GT region they share the most overlap with is the same: merging these regions would decrease $|R|$ without increasing the level of under-segmentation.
The merge network learns to solve this problem by locally operating over pairs of neighboring regions, $r_{i}$ and $r_{j}$, and making binary predictions whether or not the two regions should be merged together.
It receives a point cloud as input, where 1/4 of the points come from $r_{i}$, 1/4 of the points come from $r_{j}$, and 1/2 come from nearby outside points within a 0.1 radius (Figure \ref{fig:overview}, bottom-right).

Within the merging module, the merge network is applied in an iterative procedure.
Each round of this procedure begins by finding all neighboring regions (using region-to-region minimum point distance) and adding them into a queue to be evaluated by the merge network.
The merge network then predicts a merge probability for each pair of regions and sorts these predictions from most-to-least confident.
SHRED then iterates through these predictions, merging together neighboring regions if their merge probability is above a user-defined merge-threshold; once a region has been merged, it cannot be further merged in the same round.
The procedure repeats until a round ends with no merges.

The merge-threshold, $MT$, defines the granularity of the output decomposition. 
By default, we set $MT$ to 0.5, which encourages decompositions to match the granularity of $R^{*}$ from the training distribution. 
When set to 1, no regions are merged from the output of the fix module; when set to 0, all regions are combined together.
The merge-threshold can be used to explore tradeoffs between region granularity and region under-segmentation, as seen in Figure \ref{fig:teaser}.

\paragraph{Model Details}
We model the merge network with a PointNet++ classification back-bone, with a MLP head that predicts a binary logit.
The back-bone uses 3 set abstraction layers with 1024, 256, 64 grouping points, 0.1, 0.2, 0.4 radius size and 64, 128, 256 feature size respectively, with the global pooling step done at a feature size of 1024. 
Batch normalization and ReLU activations are used throughout. A MLP head for per-shape predictions uses 2 hidden layers with dimensions of 256 and 64. 

The input point clouds contain 2048 points, with 8 features (xyz position, normal, one hot flags for region $i$ and region $j$), and are normalized to the unit-sphere. 
The network is trained with Binary Cross Entropy loss, where 1.0 (0.0) indicates the regions should (should not) be merged together. 

\paragraph{Data Preparation}
Training data for the merge network is produced with a synthetic perturbation procedure.
This procedure begins by sampling a shape from our training set.
It then decomposes the shape into regions using our FPS procedure.
Each region in this segmentation then undergoes further decomposition in an annotation-aware fashion.
For every region, each GT part instance in that region is randomly split into sub-parts.
After this split, each sub-part is then randomly assigned to either its own region or grouped with other sub-parts in its parent region.
We empirically find this process creates decompositions that broadly cover the distribution of outputs created by the split and fix modules.

After we have created this region decomposition, we find all neighboring regions and begin sampling merges between neighbors at random.
We record each sampled merge as a training example.
Points from the two neighboring regions, $r_i$ and $r_j$, and nearby points outside both regions, form the input point cloud. 
The label of the merge is determined by checking the parts in $R^{*}$ that have the best match with the sampled regions: if $r_i$ and $r_j$ best match to the same $r^{*}_k$ part, then the merge should happen, otherwise the merge should not happen.
While sampling merges, we also update the region decomposition.
If the merge should happen, we execute the merge with 75\% chance, while if the merge should not happen, we execute the merge with 25\% chance.
This process repeats until all remaining neighbors have been considered.
\section{Experiments}
\label{sec:experiments}

We evaluate SHRED's ability to produce high-quality region decompositions using fine-grained part annotations from the PartNet dataset (Section \ref{sec:exp_srd_data}).
We describe our training procedure in Section \ref{sec:exp_srd_train}. 
In Section \ref{sec:exp_tradeoff}, we compare SHRED against related methods, analyzing the trade-off between decomposition quality and decomposition granularity.
We then examine how SHRED can be used to improve performance on downstream tasks such as zero-shot fine-grained part instance segmentation (Section \ref{sec:exp_inst}) and few-shot fine-grained semantic segmentation (Section \ref{sec:exp_sem_seg}).
Finally, we analyze the importance of SHRED's various components through a series of  ablation studies in Section \ref{sec:exp_abl}.

\subsection{Dataset Details}
\label{sec:exp_srd_data}

We use data from the finest-grained part annotations of PartNet ~\cite{PartNet} to train and evaluate our method. 
We train SHRED on a subset of categories with abundant data (in-domain), and show that SHRED is able to learn patterns that generalize well to novel categories (out-domain) at test time.
Our in-domain categories are: chair, lamp, and storage.
Our out-domain categories are: bed, display, earphone, faucet, knife, refrigerator and table.
We evenly sample shapes from our in-domain categories to form  train/validation/test splits of 6000/600/600 shapes, while each out-domain category contributes $\sim$200 shapes to the test set.

\subsection{Training Details}
\label{sec:exp_srd_train}

\begin{figure}[t!]
  \centering
  \includegraphics[width=\linewidth]{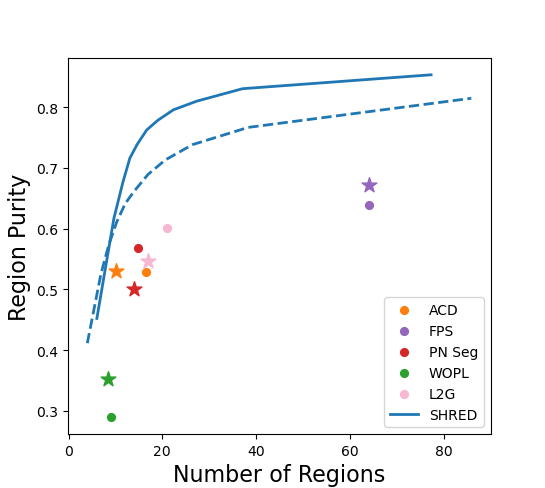}
    \caption{ Comparing segmentation granularity (X-axis, lower is better) and quality (Y-axis, higher is better). In(out)-domain averages are shown with solid (dotted) lines and circles (stars).}
    \label{fig:reg_vs_cov} 
\end{figure}

\begin{table*}[t!]
    \centering    
    \caption{Fine-grained instance segmentation performance on in-domain (left) and out domain (right) test-set shapes (metric is AIoU).
    SHRED outperforms all baseline methods, and can be further improved by setting the merge-threshold to 0.8.}
    \begin{tabular}{@{}lccccccccccccc@{}}
        & \multicolumn{4}{c}{\textbf{In Domain}} & & \multicolumn{8}{c}{\textbf{Out Domain}} \\
        \cmidrule(lr){2-5} \cmidrule(lr){7-14}
		\textbf{Method} & \textit{Avg} & \textit{Chair} & \textit{Lamp} & \textit{Storage} &  & \textit{Avg} & \textit{Bed} & \textit{Display} & \textit{Earphone} & \textit{Faucet} & \textit{Knife} & \textit{Fridge} & \textit{Table} \\
		\midrule
		FPS & 0.237 & 0.278 & 0.245 & 0.186 & & 0.173 & 0.199 & 0.115 & 0.162 & 0.202 & 0.128 & 0.151 & 0.255 \\
		WOPL & 0.178 & 0.173 & 0.248 & 0.114 &  &0.235 & 0.074 & 0.234 & 0.267 & 0.290 & 0.461 & 0.134 & 0.181 \\
		PN Seg & 0.377 & 0.381 & 0.361 & 0.389 &  & 0.318 & 0.167 & 0.427 & 0.274 & 0.306 & 0.396 & 0.253 & 0.404 \\
		L2G & 0.425 & 0.456 & 0.477 & 0.342 &  &0.392 & 0.253 & 0.466 & 0.366 & 0.451 & 0.504 & 0.247 & 0.455 \\
		ACD & 0.352 & 0.407 & 0.471 & 0.179 & & 0.393 & 0.186 & 0.481 & 0.377 & 0.469 & \textbf{0.656} & 0.158 & 0.425 \\
		SHRED ($MT = .5$) & 0.614 & 0.610 & 0.633 & 0.601 & & 0.524 & 0.426 & \textbf{0.568} & 0.408 & 0.584 & 0.606 & 0.430 & 0.644 \\
		SHRED ($MT = .8$) & \textbf{0.631} & \textbf{0.626} & \textbf{0.647} & \textbf{0.618} &  &\textbf{0.534} & \textbf{0.455} & 0.540 & \textbf{0.447} & \textbf{0.592} & 0.626 & \textbf{0.435} & \textbf{0.645} \\

        \bottomrule
    \end{tabular}
    \label{tab:inst_seg}
\end{table*}

SHRED requires training 3 networks: the split network, the fix network, and the merge network.
We use the Adam optimizer \cite{Kingma2014AdamAM} with learning rates of $1e^{-3}$, $1e^{-4}$, $1e^{-4}$ and batch sizes of 64, 64, 128 for the split, fix, and merge networks respectively.
For the merge network, a learning rate scheduler is employed to drop the learning by a factor of 0.25 every time the train loss does not reach a new minimum over a patience of 10000 iterations.
Training is performed on a machine with a GeForce RTX 3090 Ti GPU and an Intel i7-11700K CPU.
The split network trained for 70 epochs (18 hours), the fix network trained for 200 epochs (18 hours), and the merge network trained for 700k iterations ($\sim$3 days). 
All networks perform early stopping on the in-domain validation set.

\subsection{Decomposition Quality vs Granularity}
\label{sec:exp_tradeoff}

We compare SHRED against baseline methods on their ability to navigate the trade-off between decomposition quality and decomposition granularity.
Given shape $S$ with GT decomposition $R^{*}$, the objective is to produce a decomposition $R$ such that (i) $r_i \subseteq r^{*}_{j}~\forall~ i \in N $ and (ii) $|R| \leq |R^{*}|$. 
As these conditions are infeasible to meet in practice, we evaluate the goodness of a region decomposition $R$ by analyzing the degree to which it violates these properties. 
We capture violations of (i) with a region purity metric, explained below. Adherence to property (ii) can be easily captured by $|R|$, where the goal is to minimize this value. 

\paragraph{Region Purity Metric}
Given a shape $S$, region decomposition $R$ and ground-truth decomposition $R^{*}$, the region purity metric aims to capture the quality of $R$ irrespective of $R$'s granularity.
By quality we refer to the degree of under-segmentation present in $R$ w.r.t $R^{*}$, i.e. the degree to which property (i), $r_i \subseteq r^{*}_{j}~\forall~ i \in N $, is violated .

The region purity metric takes values from 0 to 1: 1 indicates (i) is not violated, while lower values indicate it has been violated to a greater degree. 
Region purity is calculated by the following procedure.
First, we find an optimal assignment $A$ from the regions of $R$ to the regions of $R^{*}$. 
$A$ is a region decomposition that keeps track of the best $r^{*}_{k} \in R^{*}$ for each $r_{i} \in R$.
For each $r_i \in R$, we calculate the ground-truth region $r^{*}_{k}$ that $r_i$ best matches:  $r^{*}_{k} = \max_{r^{*}_j \in R^{*}} IoU(r_i, r^{*}_{j})$.
Then, we find all points in $S$ assigned to $r_i$ under $R$, and set their label to $r^{*}_{k}$ under $A$.
Notice that as $R$ is a valid region decomposition, we are guaranteed that $A$ will also be a valid region decomposition.
Once $A$ has been computed, we calculate how well $A$ matches $R^{*}$.
For each ground-truth region, $r^{*}_{j} \in R^{*}$, we find all points in $S$ assigned to $r^{*}_{j}$ and calculate the percentage of those points assigned to $r^{*}_{j}$ under $A$.
The final region purity metric for the ($S$, $R$, $R^{*}$) triplet is then the average of this value across all ground-truth regions of $R^{*}$.

\paragraph{Comparisons} We evaluate SHRED against a suite of baseline methods:

\begin{itemize}
    \item \textbf{FPS}: Farthest-point sampling, the naive region decomposition that initializes SHRED.
    \item \textbf{ACD}: Approximate Convex Decomposition, a popular non-learning based method \cite{acd_pybullet}
    \item \textbf{PN Seg}: Learning-based method that predicts global fine-grained instance segmentations \cite{PartNet}.
    \item \textbf{WOPL}: Learning-based method that locally splits and globally merges regions with clustering \cite{Wang_2021_CVPR}.
    \item \textbf{L2G}: Learning-based method that uses a policy gradient trained network to locally merge regions \cite{Luo2020Learning}. 
\end{itemize}

WOPL and PN Seg use the same training data as SHRED. L2G is trained on a superset of data that SHRED uses (same in-domain categories, more shapes per category). Please see the supplemental for implementation details.

We plot the trade-off that each method makes between region quality (region purity, Y-axis) and region granularity (number of regions, X-axis) in Figure \ref{fig:reg_vs_cov}. 
The solid line and circles indicate in-domain averages, while the dotted line and stars indicate out-domain averages.
Methods that produce better region decompositions will be closer to the top-left corner. 

SHRED can vary the granularity of its output decomposition by modulating the merge-threshold (Section \ref{sec:met_merge}); we leverage this property to plot SHRED decomposition results as a curve, varying the merge-threshold from 0.01 to 0.99.
Notice that this SHRED curve forms a Pareto frontier, dominating all other comparison methods in terms of region purity (Y-axis) at any level of decomposition granularity (X-axis).
This demonstrates that, irrespective of the desired granularity, SHRED finds region decompositions that better respect the GT part regions compared with any baseline methods. 
Importantly, this trend holds on both in-domain and out-domain categories, indicating that SHRED has learned decomposition policies that generalize well to novel types of shapes. 
We provide additional results for different stages of baseline methods in the supplemental material.

\begin{figure*}[t!]
    \centering
    \setlength{\tabcolsep}{1pt}
    \begin{tabular}{ccccccccc}
       ACD & PN Seg & L2G & SHRED & GT Parts & \hspace{1.5em} & No Reg & SHRED+NGSP & GT Sem
        \\
        \includegraphics[{width=.115\linewidth}]{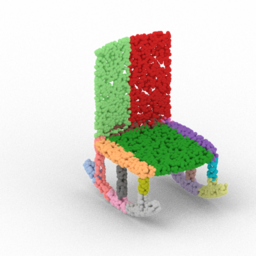} &
        \includegraphics[{width=.115\linewidth}]{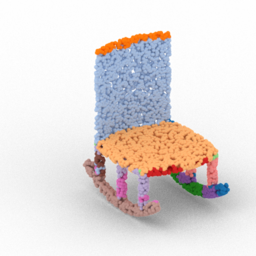} &
        \includegraphics[{width=.115\linewidth}]{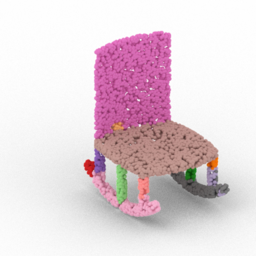} &
        \includegraphics[{width=.115\linewidth}]{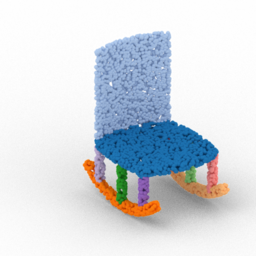} &
        \includegraphics[{width=.115\linewidth}]{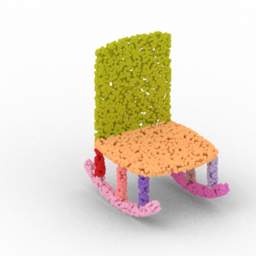} & &
        \includegraphics[{width=.115\linewidth}]{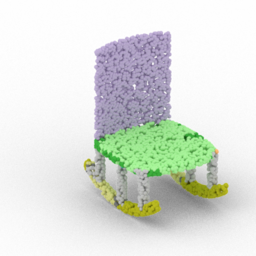} &
        \includegraphics[{width=.115\linewidth}]{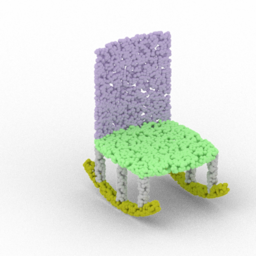} &
        \includegraphics[{width=.115\linewidth}]{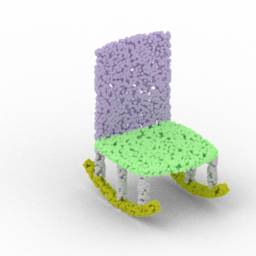} \\
        \includegraphics[{width=.115\linewidth}]{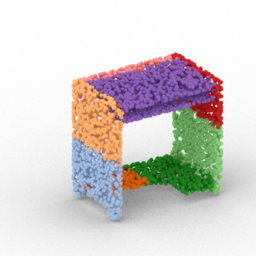} &
        \includegraphics[{width=.115\linewidth}]{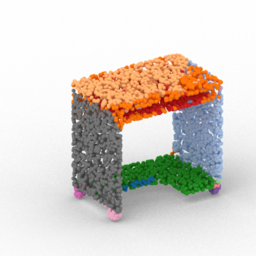} &
        \includegraphics[{width=.115\linewidth}]{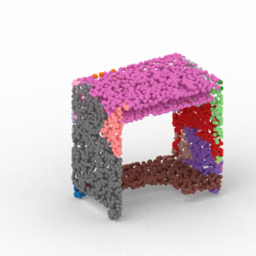} &
        \includegraphics[{width=.115\linewidth}]{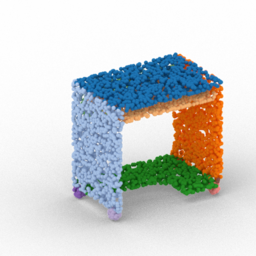} &
        \includegraphics[{width=.115\linewidth}]{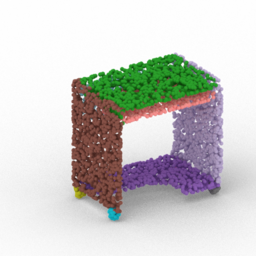} & &
        \includegraphics[{width=.115\linewidth}]{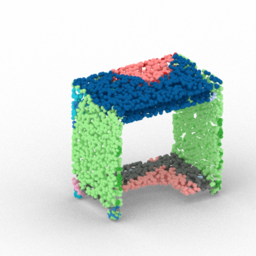} &
        \includegraphics[{width=.115\linewidth}]{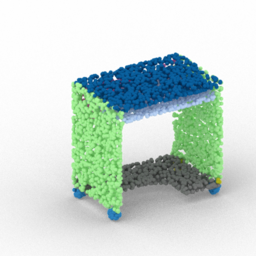} &
        \includegraphics[{width=.115\linewidth}]{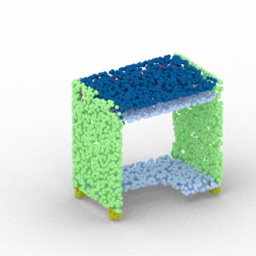}\\
        \includegraphics[{width=.115\linewidth}]{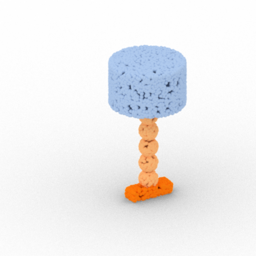} &
        \includegraphics[{width=.115\linewidth}]{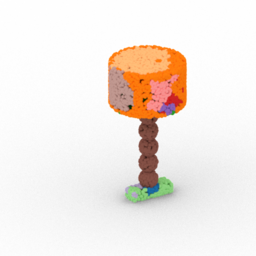} &
        \includegraphics[{width=.115\linewidth}]{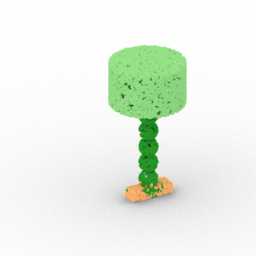} &
        \includegraphics[{width=.115\linewidth}]{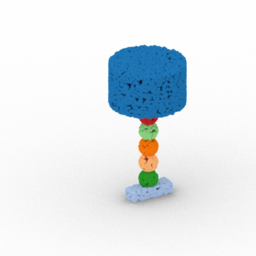}&
        \includegraphics[{width=.115\linewidth}]{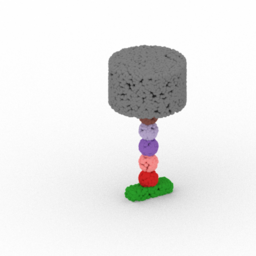}& &
        \includegraphics[{width=.115\linewidth}]{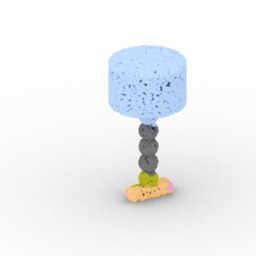} &
        \includegraphics[{width=.115\linewidth}]{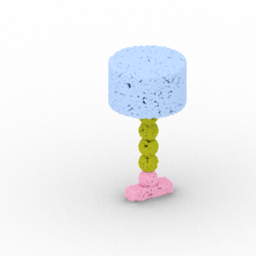} &
        \includegraphics[{width=.115\linewidth}]{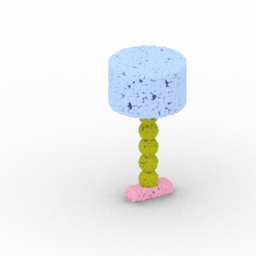} \\
        \includegraphics[{width=.115\linewidth}]{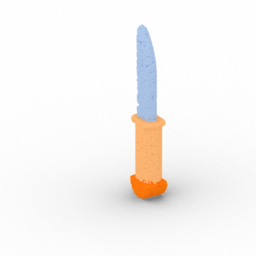} &
        \includegraphics[{width=.115\linewidth}]{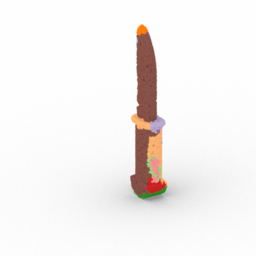} &
        \includegraphics[{width=.115\linewidth}]{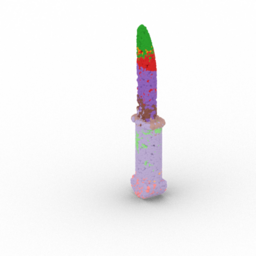} &
        \includegraphics[{width=.115\linewidth}]{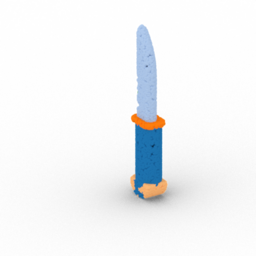} &
        \includegraphics[{width=.115\linewidth}]{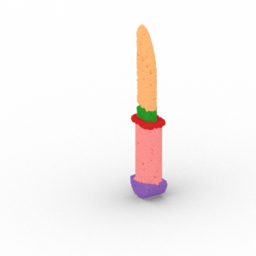} & &
        \includegraphics[{width=.115\linewidth}]{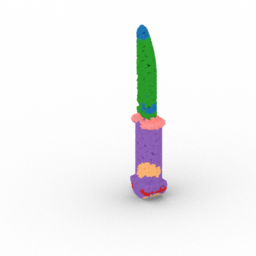} &
        \includegraphics[{width=.115\linewidth}]{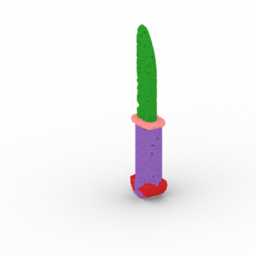}&
        \includegraphics[{width=.115\linewidth}]{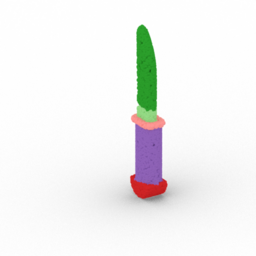} \\

    \end{tabular}
    \caption{ \emph{Left:} Qualitative region decomposition outputs, where regions are randomly colored.
    SHRED's predictions most closely match ground-truth part annotations (GT Parts). \emph{Right:} Using NGSP to label SHRED's regions creates semantic segmentations that are more similar to ground-truth semantic annotations (GT Sem) compared with region-agnostic baselines (No Reg), when labeled data is limited. }
    \label{fig:qual_seg_comp}
\end{figure*}

\subsection{Fine-grained Part Instance Segmentation}
\label{sec:exp_inst}

A straightforward application of SHRED is to treat its region decompositions as fine-grained part instance segmentations. 
We evaluate SHRED against comparison methods on this task, splitting results for in-domain test-set shapes (generalization) and out-domain test-set shapes (zero-shot).  
Following \cite{Wang_2021_CVPR}, we use AIoU as our part instance segmentation metric. 
For each GT part, we find the predicted part with the maximum IoU. 
The AIoU is then the average of this value across all GT part instances, where higher AIoU values are better.

We present quantitative results of our experiments in Table \ref{tab:inst_seg} and qualitative results in Figure \ref{fig:qual_seg_comp}, left. 
We find SHRED outperforms comparison methods on fine-grained instance segmentations for both in-domain and out-domain categories. 
For in-domain categories SHRED with a 0.5 merge-threshold achieves a 44\% boost over the next best method (L2G), while for out-domain categories SHRED provides a 33\% boost over the next best method (ACD).

\begin{figure}
  \centering
  \includegraphics[width=\linewidth]{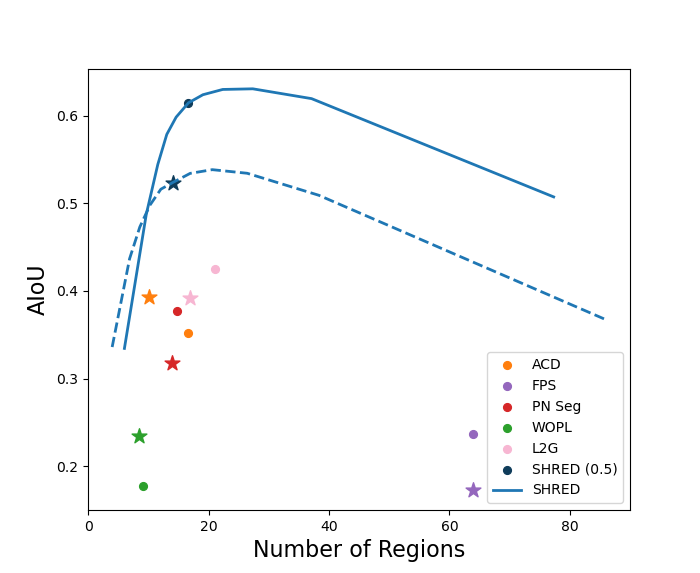}
    \caption{We plot fine-grained instance segmentation performance (AIoU) as a function of the number of predicted regions. SHRED with the default merge-threshold is shown in dark-blue, while we also vary the merge-threshold from 0.01 to 0.99 to form a curve of SHRED results (blue).}
    \label{fig:reg_vs_aiou_sup} 
\end{figure}

We analyze how the merge-threshold hyperparameter affects \\
SHRED's fine-grained instance segmentation performance.
In Figure \ref{fig:reg_vs_aiou_sup}, we plot region decomposition granularity (X-axis) versus instance segmentation AIoU (Y-axis).
As in other plots, we are able to represent SHRED results as a curve by modulating the merge-threshold from 0.01 to 0.99. 
When the merge-threshold is set to very-low or very-high values, the AIoU performance deteriorates.
We plot the performance of the default merge-threshold value, 0.5, as a blue circle (star) for the in-domain (out-domain) average. 
The default merge-threshold value achieves close to optimal performance out of all merge-threshold values, but the AIoU performance can be slightly improved by increasing the merge-threshold to 0.8 (see Table \ref{tab:inst_seg}, last row).

\begin{table}
    \centering    
    \footnotesize
    \caption{Semantic segmentation results in a few-shot paradigm (\# Train) with no regions (No Reg) and combining NGSP with region decomposition methods.
    SHRED+NGSP achieves the best mIoU performance averaged across categories.}
    \begin{tabular}{@{}llcccccc@{}}
        & & & \multicolumn{3}{c}{\textbf{In Domain}} &  \multicolumn{2}{c}{\textbf{Out Domain}} \\
        \cmidrule(lr){4-6} \cmidrule(lr){7-8}
        \textbf{\# Train} & \textbf{Method} & \textit{Avg} & \textit{Chair} & \textit{Lamp} & \textit{Storage} & \textit{Table} & \textit{Knife} \\
         \midrule
        \multirow{5}{*}{\textit{10 shapes}}
        & No Reg & 0.148 & 0.188 & 0.104 & 0.231 & 0.068 & \textbf{0.335} \\
        & PN Seg  & 0.203 &	0.190 &	0.210 &	0.287 &	0.123 &	0.075\\
        & L2G  & 0.240 &	0.276 &	\textbf{0.239}	 & 0.263 &	0.180 &	0.132 \\
        & ACD  & 0.154 &	0.193 &	0.187 &	0.115 &	0.119 &	0.151\\
        & SHRED & \textbf{0.277	} & \textbf{0.311} &	0.229	 & \textbf{0.365} &	\textbf{0.203} &	0.205\\
        \midrule
        \multirow{5}{*}{\textit{40 shapes}}
        & No Reg & 0.276 &	0.345	& 0.177	& 0.312	& 0.268	 & \textbf{0.417}\\
        & PN Seg  & 0.298 &	0.305 &	0.324 &	0.332 &	0.232 &	0.128\\
        & L2G  &  0.328	& 0.407 &	0.331	 & 0.316 &	0.259 &	0.334\\
        & ACD  & 0.237 &	0.304 &	0.302 &	0.168 &	0.174 &	0.254\\
        & SHRED & \textbf{0.375}	 & \textbf{0.431} &	\textbf{0.344} &	\textbf{0.415} &	\textbf{0.311} &	0.355\\
        \bottomrule
    \end{tabular}
    \label{tab:sem_seg}
\end{table}

\subsection{Fine-grained Semantic Segmentation}
\label{sec:exp_sem_seg}

We examine how SHRED can be used to improve fine-grained semantic segmentation performance when access to semantic label annotations are limited.
The Neurally-Guided Shape Parser (NGSP) is a method that learns to assign semantic labels to regions of a 3D shape; it has demonstrated advantages over globally learned semantic segmentation networks when it receives well-structured (e.g. human-produced) region decompositions as input \cite{NGSP}. 
We evaluate the performance of NGSP when SHRED is used to produce the input region decomposition compared with alternative methods.
For each category, NGSP requires training two types of models: a guide network and likelihood networks. 
We train a guide network separately for each region decomposition method. 
As likelihood network training is expensive, we train a single likelihood network using ground-truth part instance annotations, and share this network across experimental conditions.
Please see the supplemental material for additional experiment details.

\paragraph{Combining SHRED and NGSP} We present results of this experiment in Table \ref{tab:sem_seg} using the intersection of categories studied by our method and NGSP.
The rows in the top (bottom) half of the table correspond to training the guide and likelihood networks on 10 (40) shapes with semantic label annotations.
The metric used is semantic mIoU, where higher-values indicate a better semantic segmentation.
As seen, using SHRED produced regions allows NGSP to achieve the best semantic segmentation (last-rows) compared with using baseline methods to produce shape regions (middle-rows), or using no regions (top-row, PartNet semantic segmentation network \cite{PartNet}).
This result generally holds for both individual in-domain and out-of-domain categories.
Interestingly, for the knife category, the no region method outperforms all NGSP variants.
We attribute this result to the fact that the knife grammar from PartNet is relatively coarse and contains a top-level binary split into two sub-categories of cutting instruments (knives versus daggers) with virtually identical sub-trees.
We share qualitative results in Figure \ref{fig:qual_seg_comp}, right. 
As demonstrated, combining SHRED with NGSP typically produces fine-grained semantic labels that better match GT annotations. 
One limitation of NGSP is that it is not able to correct under-segmentations provided by region decomposition methods. For instance, the blade (dark-green) and the bolster (light-green) of the knife in the last row of Figure \ref{fig:qual_seg_comp}, are grouped into the same region by SHRED, so NGSP is unable to accurately segment out the bolster. 
To improve performance in these cases, it may be useful to run iterative rounds of SHRED and NGSP, so that SHRED can further decompose regions that NGSP believes may be under-segmented.

\paragraph{Combining SHRED and Guide} 

As NGSP guide network training is fast, we include additional semantic segmentation results in Table \ref{tab:sem_seg_guide}, where a partial version of NGSP, with just the guide network and no likelihood networks, is used to create semantic segmentations.
We include results for additional categories (all in-domain and out-domain categories studied in Section 4).
We also include additional region decomposition methods: FPS, WOPL (with just the prior network), and SHRED with different merge-threshold values (0.2, 0.5, 0.8). 
The guide network benefits most from SHRED generated regions for both in-domain and out-domain averages.
When semantic labeling grammars are more coarse, and when more labeled training shapes are present, the gap between SHRED and alternative approaches shrinks.
The chair, lamp, storage and table categories all have more than 18 nodes in their semantic grammars, while the bed, display, earphone, faucet, knife and refrigerator categories all have less than 11 nodes in their semantic grammars.

\begin{table*}
    \centering    
    \small
    \caption{Semantic segmentation mIoU performance using the NGSP guide network (no likelihood networks) to assign semantic labels to shape regions produced by different decomposition methods. We show how the guide network performs under different settings of SHRED, varying the merge-threshold from 0.2 to 0.5 to 0.8 .}
    \begin{tabular}{@{}llccccccccccccc@{}}
        & & \multicolumn{4}{c}{\textbf{In Domain}} & & \multicolumn{8}{c}{\textbf{Out Domain}} \\
        \cmidrule(lr){3-6} \cmidrule(lr){8-15}
        \textbf{\# Train} & \textbf{Method} & \textit{Avg} & \textit{Chair} & \textit{Lamp} & \textit{Storage} & & \textit{Avg} & \textit{Bed} & \textit{Display} & \textit{Earphone} & \textit{Faucet} & \textit{Knife} & \textit{Fridge} & \textit{Table} \\
         \midrule
        \multirow{9}{*}{\emph{10 shapes}}
        & No Reg & 0.174 &	0.188 &	0.103 &	0.230 & &		0.331 &	0.267 &	0.666 &	0.263 &	0.346 &	0.335 &	0.373 &	0.068 \\
        & FPS  & 0.221	 & 0.250  & 	0.205 & 	0.207 & 	 & 	0.336 & 	0.241 & 	0.607 & 	\textbf{0.274} & 	\textbf{0.390} & 	0.327 & 	0.367 & 	0.147 \\
        & WOPL Prior &  0.207 & 	0.231 & 	0.149 & 	0.241 &  & 		0.336 & 	0.259 & 	0.590 & 	0.265 & 	0.357 & 	\textbf{0.338} & 	0.362 & 	0.180 \\
        & PN SEG  &  0.213 & 	0.230 & 	0.180 & 	0.230 & 	 & 	0.285 & 	0.232 & 	0.579 & 	0.193 & 	0.263 & 	0.216 & 	0.395 & 	0.117 \\
        & L2G  & 0.233 & 	0.247 & 	0.205 & 	0.247 & 	 & 	0.293 & 	0.256 & 	0.593 & 	0.218 & 	0.262 & 	0.210 & 	0.354 & 	0.157\\
        & ACD  & 0.188 & 	0.260 & 	0.188 & 	0.117 & 	 & 	0.296 & 	0.219 & 	0.669 & 	0.218 & 	0.348 & 	0.265 & 	0.229 & 	0.127\\
        & SHRED ($MT = .2$) & 0.259 & 	0.251 & 	\textbf{0.220} & 	0.307 & 	 & 	0.315 & 	0.250 & 	0.623 & 	0.174 & 	0.322 & 	0.227 & 	0.471 & 	0.141\\
        & SHRED ($MT = .5$) & \textbf{0.279} &	0.287 &	0.193 &	\textbf{0.357} & &		0.356 &	0.303 &	\textbf{0.690} &	0.190 &	0.380 &	0.279 &	\textbf{0.494} &	0.159 \\
        & SHRED ($MT = .8$) & 0.265 & 	\textbf{0.297} & 	0.183 & 	0.316 & 	 & 	\textbf{0.360} & 	\textbf{0.309} & 	0.653 & 	0.242 & 	0.341 & 	0.328 & 	0.466 & 	\textbf{0.181}\\
        \midrule
        \multirow{9}{*}{\emph{40 shapes}}
        & No Reg & 0.278 & 	0.345 & 	0.177 & 	0.312 &  & 		0.443 & 	\textbf{0.452} & 	0.714 & 	0.371 & 	0.390 & 	\textbf{0.417} & 	0.487 & 	0.268\\
        & FPS  & 0.276 & 	0.275 & 	0.292 & 	0.261 & 	 & 	0.399 & 	0.326 & 	0.654 & 	\textbf{0.419} & 	0.436 & 	0.355 & 	0.353 & 	0.247\\
        & WOPL Prior & 0.316 & 	0.322 & 	0.312 & 	0.314 &  & 		0.416 & 	0.355 & 	0.678 & 	0.408 & 	0.415 & 	0.410 & 	0.381 & 	0.266\\
        & PN SEG  & 0.299 & 	0.314 & 	0.239 & 	0.344 &  & 		0.367 & 	0.293 & 	0.622 & 	0.331	 & 0.357 & 	0.317 & 	0.425 & 	0.226\\
        & L2G  &  0.332	 & 0.380 & 	0.302 & 	0.313 & 	 & 	0.383 & 	0.327 & 	0.679	 & 0.372 & 	0.382 & 	0.297 & 	0.397 & 	0.229\\
        & ACD  & 0.266 & 	0.328 & 	0.296 & 	0.173 &  & 		0.361 & 	0.248 & 	0.702 & 	0.356 & 	0.373 & 	0.412 & 	0.248 & 	0.190\\
        & SHRED ($MT = .2$) & 0.343 & 	0.365 & 	0.298 & 	0.367 & 	 & 	0.421 & 	0.395 & 	0.709 & 	0.298 & 	0.390 & 	0.338 & 	0.510 & 	0.304\\
        & SHRED ($MT = .5$) & \textbf{0.368} & \textbf{	0.399} & 	\textbf{0.325} & 	0.379 & 	 & 	0.443 & 	0.409 & 	0.722 & 	0.344 & 	0.418 & 	0.396 & 	\textbf{0.522} & 	0.287\\
        & SHRED ($MT = .8$) & \textbf{0.368} & 	0.398 & 	0.300 & 	\textbf{0.407} &  & 	\textbf{	0.465} & 	0.408 & 	\textbf{0.736} & 	0.405 & 	\textbf{0.468} & 	0.400 & 	0.504 & 	\textbf{0.331}\\
        \bottomrule
    \end{tabular}
    \label{tab:sem_seg_guide}
\end{table*}

\subsection{SHRED Ablations}
\label{sec:exp_abl}

\definecolor{gold}{RGB}{201, 176, 55}
\definecolor{silver}{RGB}{205, 205, 205}
\definecolor{bronze}{RGB}{205, 127, 50}

\begin{table}
    \centering    
    \caption{SHRED instance segmentation performance under different hyperparameter settings and ablation conditions. Metric is AIoU, averaged for both the in-domain and out-domain categories. \colorbox{gold}{Best}, \colorbox{silver}{second-best}, and \colorbox{bronze}{third-best} conditions are highlighted.}
    \begin{tabular}{@{}lcc@{}}
        \midrule
        \textbf{Condition} & \textbf{In Domain AIoU}  & \textbf{Out Domain AIoU} \\
        \midrule
		No Split & 0.470 & 0.440 \\
		No Fix & 0.574 & 0.492 \\
		No Merge & 0.324 & 0.225 \\
		\midrule
		No Hung OS match & \colorbox{bronze}{0.599} & \colorbox{bronze}{0.515} \\
			\midrule
		Chair only & 0.484 & 0.485 \\
		Lamp only & 0.329 & 0.387 \\
		Storage only & 0.444 & 0.414 \\
		Limited data (10\%) & 0.543 & 0.496 \\
			\midrule
		Split with naive SDC & 0.490 & 0.448 \\
		Align with naive SDC & 0.561 & 0.503 \\
		Merge with naive SDC  & 0.488 & 0.425 \\
			\midrule
		Cascade training & 0.434 & 0.413 \\
			\midrule
			SHRED ($K=32$) & 0.594 & 0.512 \\
			SHRED ($K=64$) & \colorbox{silver}{0.614} & \colorbox{silver}{0.524} \\
            SHRED ($K=128$)& \colorbox{gold}{0.617} & \colorbox{gold}{0.541} \\
        \bottomrule
    \end{tabular}
    \label{tab:shred_abl}
\end{table}

In this section, we consider the performance of SHRED under different hyperparameter settings and ablations conditions. We quantitatively evaluate each modified version in Table \ref{tab:shred_abl}, by measuring the achieved fine-grained instance segmentation AIoU over in-domain and out-domain category averages. In the rest of this section, we describe the different ablation conditions that populate the rows of the table.

\paragraph{Removing local operations}

SHRED uses three locally learned modules to perform region splits, fixes and merges. 
We evaluate ablated version of SHRED, where only two modules are employed to produce region decompositions, in the \textit{No Split}, \textit{No Fix}, and \textit{No Merge} rows.
As demonstrated, removing any of SHRED's local operations leads to worse region decompositions. 

\paragraph{Modified Hungarian Matching Algorithm}

As discussed in Section \ref{sec:met_split}, split network training requires a dynamically computed matching between predicted and target instance slots.
SHRED uses a variant of the typical Hungarian matching algorithm for this procedure, where over-segmentation is explicitly encouraged. 
In the \textit{No Hung OS match} row, we ablate this design decision, replacing our over-segmentation biased Hungarian matching algorithm with the typical exact instance segmentation formulation.
As seen, changing to the default formulation slightly decreases SHRED's performance.
Please see the supplemental material for a detailed explanation of our matching algorithm.

\paragraph{Training with less data}

By default, the local operators of SHRED are trained over multiple in-domain shape categories (chairs, lamps, and storage furniture) with each category contributing 2000 shape instances. 
In the \textit{Chair only}, \textit{Lamp only}, and \textit{Storage only} ablation rows, we evaluate how SHRED is able to generalize when trained over 2000 instances from a single shape category. 
In the \textit{Limited data} row, we present a version of SHRED that is trained over the same shape categories (chairs, lamps and storage furniture), but where each category contributes only 200 shape instances. 
From these results, we can observe that SHRED benefits from both (a) more in-domain shape categories and (b) more shape instances per category, but (a) should be prioritized over (b). 
In fact, SHRED with just 10\% of the training data still significantly outperforms all of the baseline methods from Table \ref{tab:inst_seg}. 

\paragraph{Naive Synthetic Data Creation}

As described throughout Section \ref{sec:method}, SHRED employs synthetic data creation (SDC) strategies to produce training data for each local operator. 
We designed these SDC strategies to roughly match the types of input each operator might expect during the course of the split-fix-merge sequence. 
In the \textit{Split with naive SDC}, \textit{Align with naive SDC}, and \textit{Merge with naive SDC} rows we ablate our SDC procedure for one operator at a time, by replacing the default SDC with a naive SDC. 
Please see the supplemental material for details on the naive synthetic data creation procedures.
In all cases, replacing the default SDC with the naive SDC leads to worse AIoU. 
This supports our claim that having a synthetic data creation strategy that broadly matches the expected input for each local operator improves SHRED's overall performance.

\paragraph{Cascade Training of Operators}

SHRED trains each operator independently with synthetic data creation procedures. 
When a single operator sequence is desired, it is also possible to train the operators of SHRED in a cascading fashion: taking the predictions from previous stages to produce the training data for later stages. 
We present a version of SHRED trained within this paradigm in the \textit{Cascade training} row.
In this paradigm,  the fix network is trained on predictions from the split network, and the merge network is trained on predictions from the split and fix networks. 
The performance of this condition is much worse compared with the default version of SHRED, likely because in the cascade approach the amount of operator training data is bottle-necked by the number of training shapes, whereas the synthetic data creation procedure can generate infinite amounts of operator training data, even with limited training shapes. 
Further note that training over synthetic data allows operators to be trained in parallel and provides greater flexibility in how the operators can be applied during an inference procedure.

\paragraph{Initial Region Splits}

SHRED applies its learned local operators over regions produced by a naive decomposition method. 
By default, we use a farthest-point sampling (FPS) procedure to produce the naive decomposition with $K=64$ centroids. 
In the last rows of Table \ref{tab:shred_abl} we evaluate SHRED's instance segmentation performance under different values of $K$.
SHRED is not overly sensitive to the initial decomposition granularity, performing well under all $K$ values.
The best AIoU value, for both in-domain and out-domain categories, is achieved with $K=128$. 
The downside of starting with a finer initial region decomposition is that SHRED inference is slower; SHRED inference with $K=64$ takes under 6 seconds per shape, while SHRED inference with $K=128$ takes over 10 seconds per shape.

\section{Conclusion}

We introduced SHRED, a method that performs 3D SHape REgion Decomposition by learning to locally split, fix and merge. 
We trained SHRED on part annotations from three data-abundant PartNet categories (chairs, lamps, storage) and experimentally validated its ability to produce high-quality region decompositions for out-domain categories of manufactured shapes. 
In comparisons against baseline methods, we used SHRED's merge-threshold hyperparameter to demonstrate that it offers the best trade-off between decomposition quality and granularity.
Finally, we evaluated SHRED on downstream applications involving fine-grained parts, zero-shot instance segmentation and few-shot semantic labeling, finding that SHRED improves performance over baselines.

\paragraph{Future Work} 

While SHRED has demonstrated an impressive capacity to generalize to out-of-distribution instances, all of the shapes we have experimented with come from the same meta-distribution (manufactured objects). 
It would be interesting to investigate how SHRED generalizes to more diverse domains, including those that share locally similar properties (partial shape scans, 3D scenes) or those with hardly any similarities (organic bodies). 
Along another direction, while SHRED's merge-threshold allows some exploration over a range of decomposition granularities, the segmentations that SHRED produces are always flat. 
Developing a procedure to convert this series of flat decompositions into a shared hierarchical segmentation may prove useful for various downstream applications (e.g. collision detection or structure-based generative modeling). 

Finally, while we find that SHRED's sequential application of the split, fix and merge operations results in region decompositions that outperform previous approaches, it is unlikely that this sequence is optimal for every shape instance.
It is easy to imagine that some input shapes might benefit from repeated applications of a particular operation, by running this iterative procedure multiple times, or by only applying the operation to certain regions.
Wrapping the operations of SHRED with a more advanced outer-loop search would be one way to accomplish this per-shape tailoring goal.
Moreover, if this search was guided by a global likelihood function, then application-dependant terms could even be added into consideration (e.g. encouraging convex regions or respecting bilateral symmetries). 
Alternatively, a human-in-the-loop could provide constraints to guide the decomposition process in an interactive fashion, by for instance, indicating patches to be split or merged, or boundaries to be fixed, with sketched-based controls.
We believe that these paradigms, where human-specified objectives help guide local data-driven modules, are a promising way forward towards generating human-quality fine-grained region decompositions of 3D objects from distributions that lack abundant annotations. 

\begin{acks}
We would like to thank the anonymous reviewers for their helpful suggestions. This work was funded in parts by NSF award \#1941808 and a Brown University Presidential Fellowship. Daniel Ritchie is an advisor to Geopipe and owns equity in the company. Geopipe is a start-up that is developing 3D technology to build immersive virtual copies of the real world with applications in various fields, including games and architecture
\end{acks}

\bibliographystyle{ACM-Reference-Format}
\bibliography{main}

\appendix

\section{SHRED Implementation Details}

\subsection{Synthetic Data Creation}
\label{sec:sdc}
\paragraph{Fix Network} To generate an input-output training example for the fix network, we employ the following procedure.
From our training dataset, we first sample a random region from a random shape, using the ground-truth fine-grained part annotations to define a region.
We increase the size of the region by Y\% (Y between 5 and 25) with a 75\% chance: this involves sampling a random point about the regions center, finding the Y\% closest points currently outside the region, and flipping their sign. 
By a similar procedure we randomly remove between 10 and 50\% of the inside region points with a 75\% chance.
Next we up or down sample the inside region points, and nearby outside points (within a 0.1 extended radius of the region's center + radius) to form two 2048 point clouds. 
Each point in this point cloud has its label flipped with some Z\% (Z randomly set to be between 0.0 and 0.3). 
To offset data imbalance that causes more inside labels to flip to outside, we finally randomly flip any extra inside points (those down-sampled in the previous step) to become outside points with a 10\% probability.
This allows the network to see more cases where outside points become inside points during training. 
To decide whether to keep or reject the entire sample, we calculate the percentage of input inside points that have inside as their target label, and the percentage of inside label points that are associated with an inside input point: if either of these percentages is below 40\%, we skip the example. 

\paragraph{Merge Network} To generate training data for the merge network, we employ the following procedure. 
First we sample a random shape from the training dataset. 
Then we use FPS to split the shape into M regions (M randomly chosen from 16, 32, 64, 128). 
For each region, we first identify all ground-truth part instances that appear in the region.
We split each of these regions into K sub-regions, where K is sampled between 1 and 10 according the probability distribution proportional to $0.5^{K}$. 
To execute this split, we sample K points randomly distributed about the center of the region, and then assign each point in the region to the randomly sampled point it is closest to. 
Each sub-part is then assigned to some cluster of regions randomly: with 1/3 chance it is merged into a default group with other sub-parts in the same FPS defined region, with 1/3 chance it is merged into one of K groups with other sub-parts in the same FPS defined region, and with 1/3 chance it is merged into one of K groups with other sub-parts in the same FPS defined region from the same GT part instance. 

At this point, we are ready to start gathering merge examples.
We first find all neighboring regions from our above synthetic split procedure.
We then randomly sample a pair of neighboring regions, and check which ground-truth parts they most heavily overlap with. 
If they overlap most with the same ground-truth part, then we record the pair as an example where a split \textit{should} happen, otherwise we record the pair as an example where a split \textit{should not} happen.
To produce merge examples at different granularities, we also modify the regions during this procedure; every time a merge should happen, we actually perform the merge 75\% of the time, and every time a merge should not happen, we actually perform the merge 25\% of the time. 
We continue collecting merge examples until we run out of neighbor pairs that have not been previously considered.

\section{SHRED Method Ablations}

\subsection{Modified Hungarian Matching Algorithm}

The split network is trained with a cross entropy loss between predicted instances and target instances.
Unlike most settings where cross entropy is used, for the instance segmentation task there is not a canonical mapping from prediction slots to target slots. 
For this reason, a typical approach has been to dynamically use the Hungarian matching algorithm to find a one-to-one mapping between prediction slots and target slots that would minimize the cross entropy loss. For an example, see \cite{PartNet}. 

While this matching scheme is optimal when an exact instance-segmentation is desired, we would prefer that the split network removes \textit{all} under-segmentation, even if this comes at the cost of producing more regions. As our merge module will later figure out how to combine different over-segments, we want the split network to solve an over-instance segmentation task.
To this end, we modify the matching procedure between prediction slots and target slots to greedily search for opportunities to reward network over-segmentation predictions.

The procedure takes as input a logit prediction over N points with K slots $P$, a $(N~x~K)$ matrix, and a ground-truth instance segmentation $T$, a $(N~x~K)$ matrix where each column is a one-hot vector.
To begin, we employ the typical Hungarian matching approach to find a one-to-one assignment $M$ between prediction and target slots.
We identify all unassigned prediction and target slots (e.g. those that correspond to empty instances).
Then for each paired unused prediction slot, $U_P$, and unused target slot, $U_T$, we find all indices in $P$ where $U_P$ was the argmax prediction.
We then index into $T$ with these indices, and find the mode region in T that best aligns with $U_P$, call this region $A$.
Next, we find the slot of $P$ that was assigned to $A$ under $M$, call this slot $P_A$.
We find the index rows of $T$ where $A$ is set to 1.0, call these $T_A$.
Then we use the logit predictions of $P_A$ and $U_P$ to split $T_A$ into two parts, $T_{PA}$ and $T_{UP}$, with an argmax operation.

At this point we have identified a possible over-segmentation: we could modify $T_A$ to become split into two parts, $T_{PA}$ and $T_{UP}$, where prediction slot $P_A$ would be re-assigned to the region $T_{PA}$ in slot $A$ and unused prediction slot $U_P$ would be assigned to region $T_{UP}$ in slot $U_T$. 
If $T_{PA}$ and $T_{UP}$ both contain more than 10 indices each, and $A$ was the argmax prediction for more than 50\% of the indices under $U_P$, then we accept this over-segmentation with modifications to $T$ and $M$. This procedure is repeated for ever pair of unused slot.

\subsection{Naive Synthetic Data Creation}

In Section 4.6 of the main paper, we present results of SHRED when learned local operators are replaced with versions trained under naive synthetic data creation (SDC) procedures. In this section, we describe the naive SDC procedures used for the split, align and merge modules for each ablation condition.

\paragraph{Naive split SDC}

To generate naive training data for the split network we employ the following procedure. Given a training shape, we first generate a region decomposition by iteratively: (1) sampling a random point from the shape, (2) sampling a random radius r, from a beta distribution with alpha as 1.5 and beta as 4.0, (3) assigning all points within r radius of the randomly sampled point to a new region (with a limit of up to half of the remaining unassigned points). If we want to produce K regions, we run this iterative procedure K-1 times, then throw all unassigned points into the Kth region (we set K=64). Once a region decomposition has been produced, we follow the logic of the default SDC, where each region in the naive decomposition contributes one input point cloud for training, and ground truth target instances are supplied by the fine-grained annotated labels.

\paragraph{Naive fix SDC}

To generate naive training data for the fix network we employ the following procedure. From our training dataset, we first sample a random surface point from a random shape. Then we sample a random radius r from the same distribution as in the Naive split SDC case. All points with r radius of the randomly sampled point are then set to be within the target region (up to 25\% of the points in the shape). Following the default procedure, we then up or down sample the inside region points, and nearby outside points (within a 0.1 extended radius of the region’s center + radius) to form two 2048 point clouds.
We then randomly flip the label of each point, with a probability of 15\%. The target inside-outside values of this example are then computed by finding the region in the ground-truth labeling that has the maximum overlap with the inside points of the example. 

\paragraph{Naive merge SDC}

To generate naive training data for the merge network we employ the following procedure. First a naive region decomposition is produced in the same manner as described in the \textit{Naive split SDC} paragraph (where K is randomly chosen from the set of 16, 32, 64, 128). Then, we generate merge examples following the default SDC logic, sampling random merges, recording whether they should happen with respect to the ground-truth region annotations, and randomly actualizing the merge (full details in the last paragraph of Section \ref{sec:sdc}).

\section{Baseline Implementation Details}

\begin{figure}
  \centering
  \includegraphics[width=\linewidth]{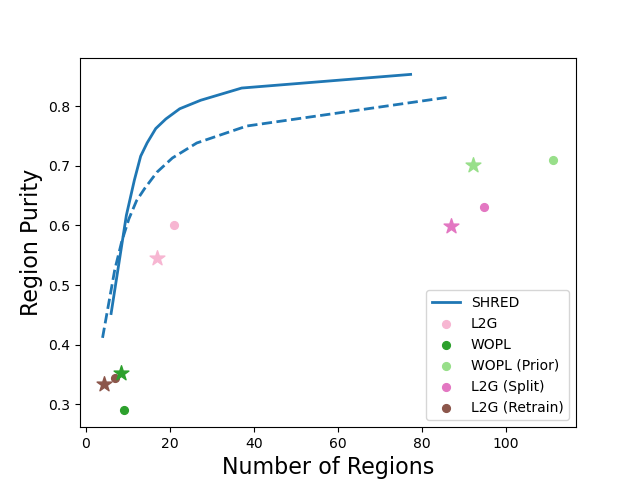}
    \caption{Trade-off between region granularity and region quality for additional baseline conditions. WOPL and L2G contain intermediary stages that correspond with SHRED's split stage, so we plot the results of these stages as separate conditions (WOPL Prior and L2G Split). Our attempt to retrain L2G on our training dataset failed to converge to good performance (brown dot).}
    \label{fig:reg_vs_cov_sup_met} 
\end{figure}

\subsection{ACD}

We use the VHACD implementation from \cite{acd_pybullet}.
This implementation is widely used by a multitude of applications, and has been used by related work to formulate self-supervised training objectives \cite{selfsupacd}. 
We found this ACD implementation works best when input meshes are manifold; as such we do not directly apply it to PartNet meshes, as many are double-sided or contain inconsistent normal orientations. 
Instead, we first send each PartNet mesh through a manifold procedure \cite{huang2018robust}, and run the VHACD algorithm on the manifold scipt output. 
ACD generated convexes are then used to partition a high-resolution point cloud (100k points) into segments by sampling the surface of each convex and finding the nearest neighbor from each point in the high-resolution point cloud to the point cloud of convexes. 

\subsection{PN Seg}

We retrain Partnet's instance segmentation model using our training data and the author's publicly released code at \\
https://github.com/daerduoCarey/partnet\_seg\_exps .

Following \cite{Luo2020Learning}, we remove the semantic label prediction head and semantic label loss, as we train over multiple categories at once.
We follow all other default hyper-parameters as found in the code.
We write a procedure to export our training data into the h5 file format that their method expects, and then start training runs through their released script.

\subsection{L2G}

L2G trains a version of their method on the same in-domain categories that SHRED uses, but uses slightly more data for each category \cite{Luo2020Learning}. 
For all experiments in the main paper, we use the author's released models from \\
https://github.com/tiangeluo/Learning-to-Group . While this causes the training distribution between SHRED and L2G to differ slightly, if anything we believe this hurts SHRED, as it has access to less data.
The L2G method has an initial step before merging where a simple model produces a shape over-segmentation; we show how this stage fairs on the trade-off graph between decomposition granularity and purity in Figure \ref{fig:reg_vs_cov_sup_met} (dark pink points).  

Using the released code, we attempted to retrain a version of L2G on the subset of chair, lamp, and storage data used to train SHRED. 
We converted our data into the same h5 file as we used for PN Seg, and then initiated L2G training with their specified scripts, keeping all method hyper-parameters unchanged.
Interestingly, we found that this version of L2G converged to much worse performance, we plot this retrained version of L2G on Figure \ref{fig:reg_vs_cov_sup_met} (brown points).

\subsection{WOPL}

The WOPL method has no publicly available code and we were unable to gain access to an implementation by contacting the authors. 
To this end, we attempted to implement the algorithm as described in the paper \cite{Wang_2021_CVPR}.
We include our implementation of the WOPL method in the released code, where in all cases we tried to follow the paper's description as closely as possible.
The WOPL method has an analogous stage to SHRED's split stage, termed the prior stage, that creates a region decomposition before a global merge step. 
In Figure \ref{fig:reg_vs_cov_sup_met} we show the trade-off the WOPL prior makes between decomposition granularity and quality (light green points).

\section{Semantic Segmentation Experiment Details}

Our experiments in Section 4.5 demonstrate how SHRED can benefit fine-grained semantic segmentation methods when training data with semantic labels is limited.
We use SHRED to generate region decompositions that are passed into NGSP: an approach that learns to assign semantic labels from a fine-grained grammar to regions of a 3D shape. 

We evaluate SHRED (and other region decomposition approaches, see PN Seg, L2G, and ACD rows of Table 2 in the main paper) under two limited labeled data paradigms: when 10 or 40 shapes with semantic labels are provided to train a semantic segmentation method. 
We first use the region decomposition methods (trained on all in-domain categories as described in sections 4.1 and 4.2) to find a region decomposition for the 10 or 40 training shapes that have semantic label annotations.
With a training dataset of these shapes that contain both semantic annotations and region decompositions, we then have the data needed to train NGSP for a given category.

NGSP trains separate networks for each category, as semantic labeling is specific to a category.
We follow NGSP's procedure to train a guide network and likelihood networks for each category \cite{NGSP}.
For our experiments in Table 2 of the main paper, we use the 5 categories studied by both SHRED and NGSP: chairs, lamps, storage, tables, and knives. 
Guide network training is relatively inexpensive.
A guide network can be trained in less than 10 minutes when the number of labeled shapes is low (the settings we are interested in), so we retrain a separate guide network for each category and for each region decomposition method (e.g PN Seg, L2G, ACD, SHRED). 
Likelihood network training is more involved, as it requires training multiple networks for each node in a fine-grained semantic grammar (as reference, the chair grammar has over 30 nodes). 
Therefore, we train a single collection of likelihood networks that are shared by all region decomposition methods; the likelihood networks use ground-truth region annotations during training (as we assume each shape with semantic annotations also has part instance annotations).

Once the guide and likelihood networks have been trained, we populate the values of Table 2 with the following procedure. 
A input shape for category C is first decomposed into a set of regions with region decomposition method M.
The guide network specialized for the (C, M) pair is then used to propose a set of 10000 label assignments over the region decomposition.
The likelihood network specialized for C then evaluates all proposed label assignments, and chooses the label assignment that maximizes equation 1 from the NGSP paper. 
Note that we also include the guide network likelihood into this equation, finding it regularizes the predictions of the likelihood networks that have never seen region decompositions produced by M. Per-region semantic label predictions can then be propagated to the points within each region using the chosen label assignment.
Finally, the semantic mIoU can be calculated by finding the intersection over union between predicted and ground-truth per-point labels, and averaging this value over the labels of the grammar across the test-set shapes of each category.

The No Reg rows in Table 2 and 3 in the main paper correspond with predictions from the fine-grained semantic segmentation network proposed by \cite{PartNet}. 
This network operates globally, and learns to label individual points instead of shape regions.

\section{Additional Qualitative Results}

We provide additional qualitative comparisons between the region decompositions produced by various methods in Figure \ref{fig:qual_seg_comp}. Columns 6-8 show how SHRED's predictions change as the merge-threshold hyper-parameter is modified (we look at the values of 0.2, 0.5, and 0.8).

  \begin{figure*}[t!]
    \centering
    \setlength{\tabcolsep}{1pt}
    \begin{tabular}{cccccccccc}
       FPS & WOPL & ACD & PN SEG & L2G & SHRED (.2) & SHRED (.5) & SHRED (.8) & GT Parts  \\
        \includegraphics[{width=.101\linewidth}]{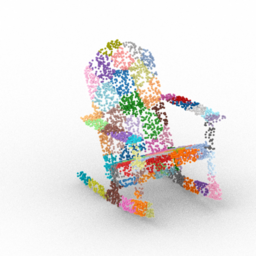} &
        \includegraphics[{width=.101\linewidth}]{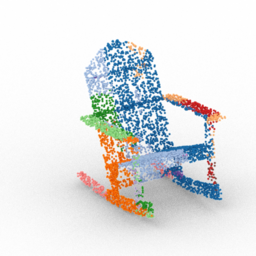} &
        \includegraphics[{width=.101\linewidth}]{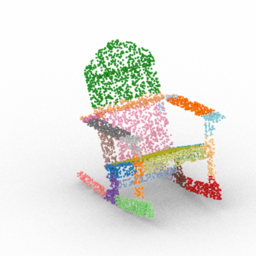} &
        \includegraphics[{width=.101\linewidth}]{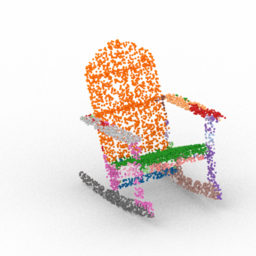} &
        \includegraphics[{width=.101\linewidth}]{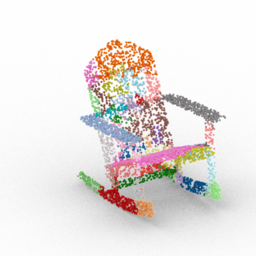}&
        \includegraphics[{width=.101\linewidth}]{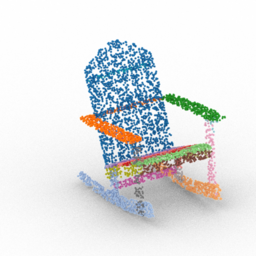} &
        \includegraphics[{width=.101\linewidth}]{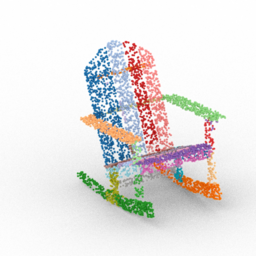} &
        \includegraphics[{width=.101\linewidth}]{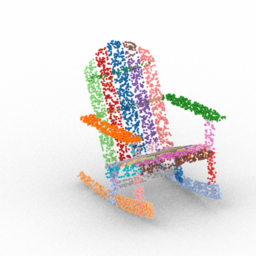} &
        \includegraphics[{width=.101\linewidth}]{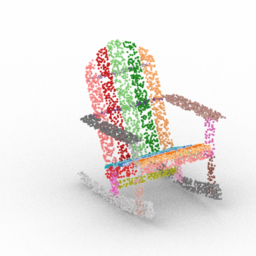} \\
                \includegraphics[trim={2.0cm 0.0cm 2.0cm 3.5cm},clip,{width=.101\linewidth}]{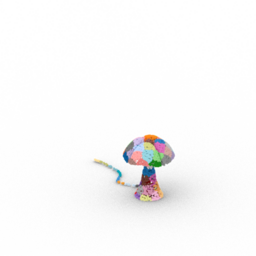}&
        \includegraphics[trim={2.0cm 0.0cm 2.0cm 3.5cm},clip,{width=.101\linewidth}]{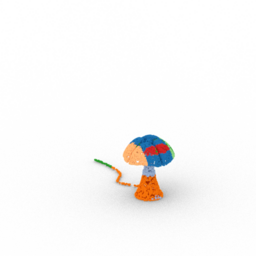} &
        \includegraphics[trim={2.0cm 0.0cm 2.0cm 3.5cm},clip,{width=.101\linewidth}]{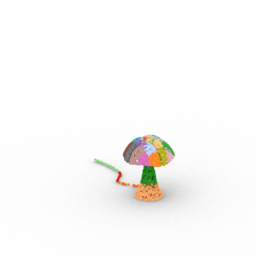} &
        \includegraphics[trim={2.0cm 0.0cm 2.0cm 3.5cm},clip,{width=.101\linewidth}]{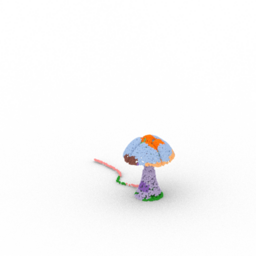} &
        \includegraphics[trim={2.0cm 0.0cm 2.0cm 3.5cm},clip,{width=.101\linewidth}]{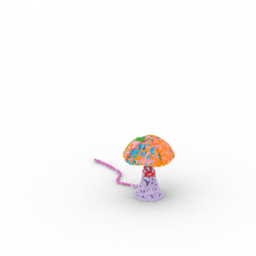}&
        \includegraphics[trim={2.0cm 0.0cm 2.0cm 3.5cm},clip,{width=.101\linewidth}]{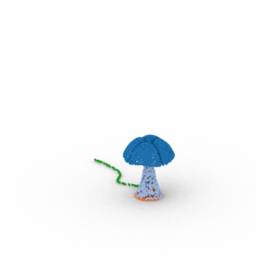} &
        \includegraphics[trim={2.0cm 0.0cm 2.0cm 3.5cm},clip,{width=.101\linewidth}]{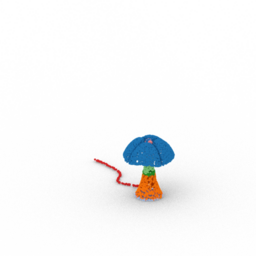}&
        \includegraphics[trim={2.0cm 0.0cm 2.0cm 3.5cm},clip,{width=.101\linewidth}]{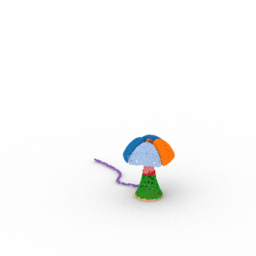} &
        \includegraphics[{trim={2.0cm 0.0cm 2.0cm 3.5cm},clip,width=.101\linewidth}]{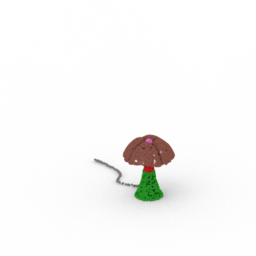}\\
                \includegraphics[{width=.101\linewidth}]{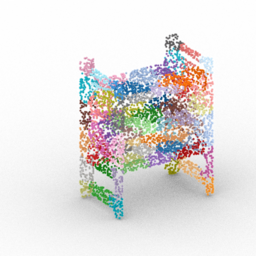} &
        \includegraphics[{width=.101\linewidth}]{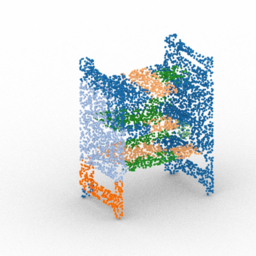} &
        \includegraphics[{width=.101\linewidth}]{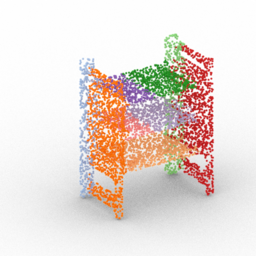}  &
        \includegraphics[{width=.101\linewidth}]{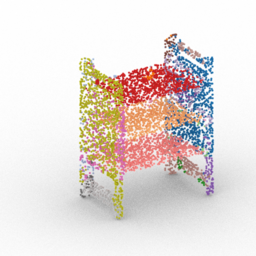}  &
        \includegraphics[{width=.101\linewidth}]{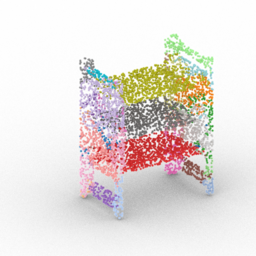}  &
        \includegraphics[{width=.101\linewidth}]{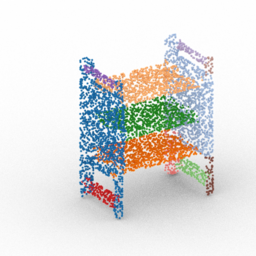}  &
        \includegraphics[{width=.101\linewidth}]{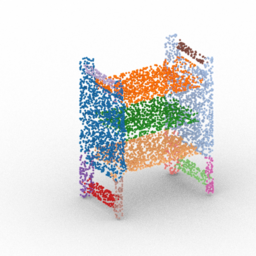}  &
        \includegraphics[{width=.101\linewidth}]{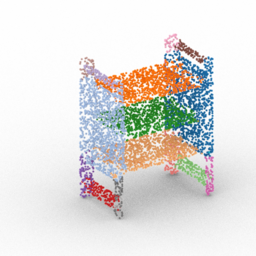}  &
        \includegraphics[{width=.101\linewidth}]{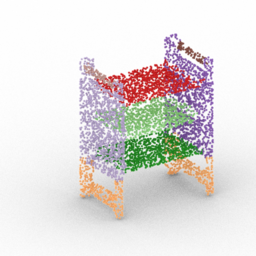} \\
                \includegraphics[{width=.101\linewidth}]{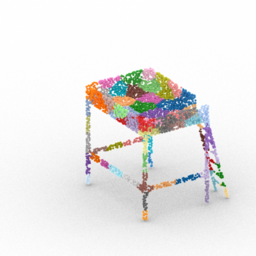}  &
        \includegraphics[{width=.101\linewidth}]{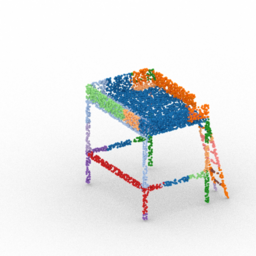} &
        \includegraphics[{width=.101\linewidth}]{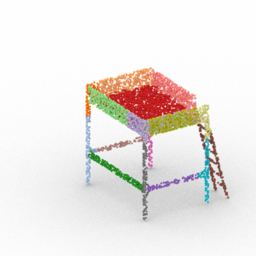} &
        \includegraphics[{width=.101\linewidth}]{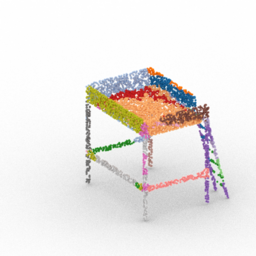} &
        \includegraphics[{width=.101\linewidth}]{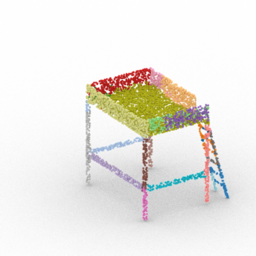} &
        \includegraphics[{width=.101\linewidth}]{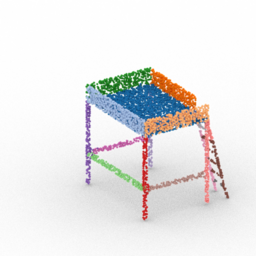} &
        \includegraphics[{width=.101\linewidth}]{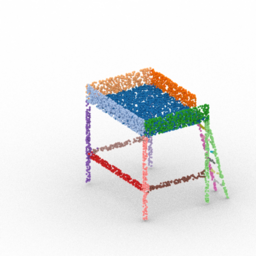} &
        \includegraphics[{width=.101\linewidth}]{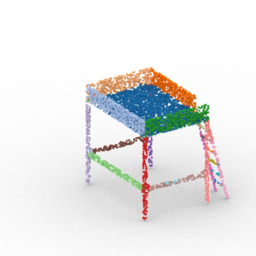} &
        \includegraphics[{width=.101\linewidth}]{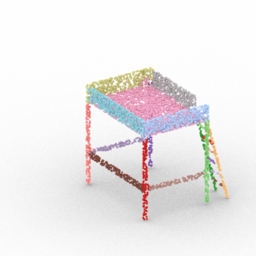}\\
                \includegraphics[{width=.101\linewidth}]{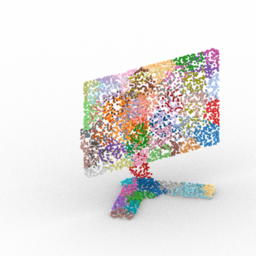} &
        \includegraphics[{width=.101\linewidth}]{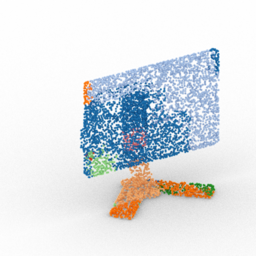} &
        \includegraphics[{width=.101\linewidth}]{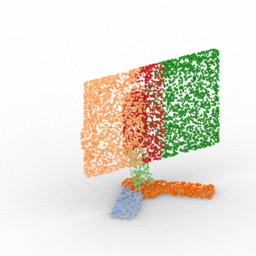} &
        \includegraphics[{width=.101\linewidth}]{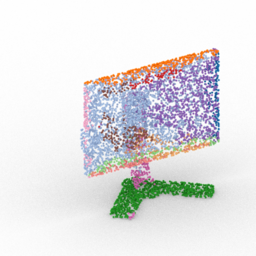} &
        \includegraphics[{width=.101\linewidth}]{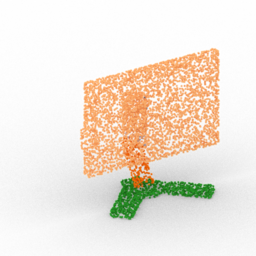} &
        \includegraphics[{width=.101\linewidth}]{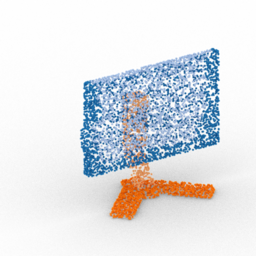} &
        \includegraphics[{width=.101\linewidth}]{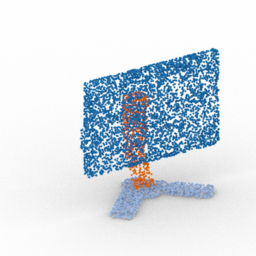} &
        \includegraphics[{width=.101\linewidth}]{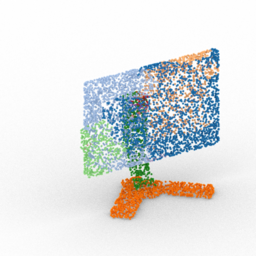} &
        \includegraphics[{width=.101\linewidth}]{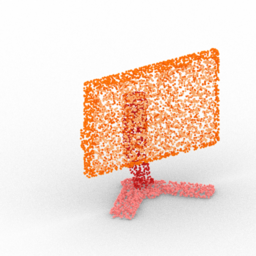}\\
                \includegraphics[{width=.101\linewidth}]{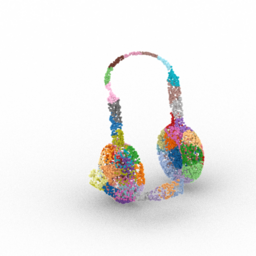} &
        \includegraphics[{width=.101\linewidth}]{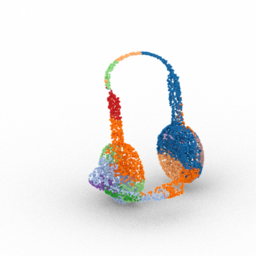}  &
        \includegraphics[{width=.101\linewidth}]{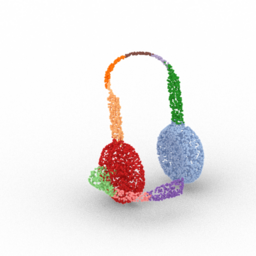}  &
        \includegraphics[{width=.101\linewidth}]{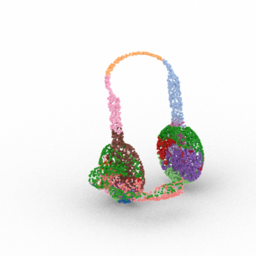}  &
        \includegraphics[{width=.101\linewidth}]{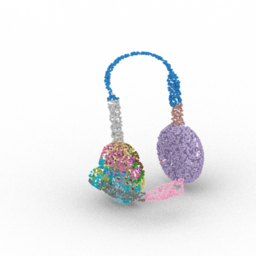}  &
        \includegraphics[{width=.101\linewidth}]{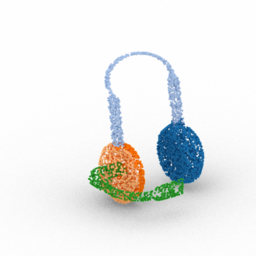}  &
        \includegraphics[{width=.101\linewidth}]{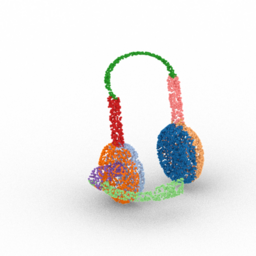}  &
        \includegraphics[{width=.101\linewidth}]{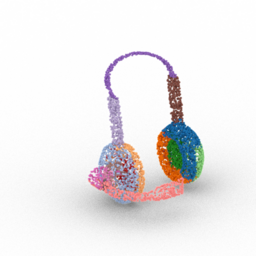}  &
        \includegraphics[{width=.101\linewidth}]{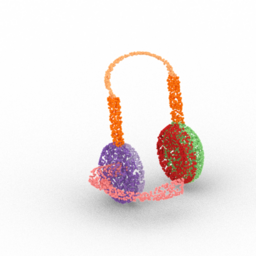} \\
                \includegraphics[{width=.101\linewidth}]{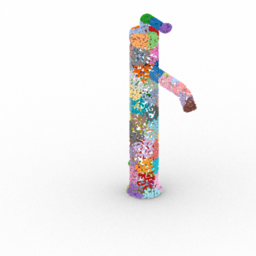}  &
        \includegraphics[{width=.101\linewidth}]{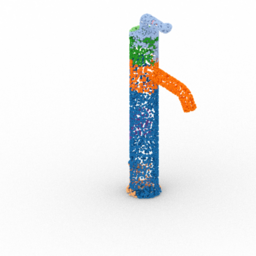}  &
        \includegraphics[{width=.101\linewidth}]{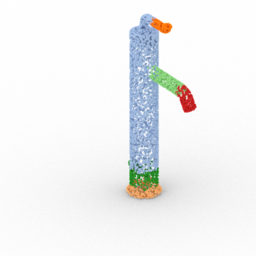}  &
        \includegraphics[{width=.101\linewidth}]{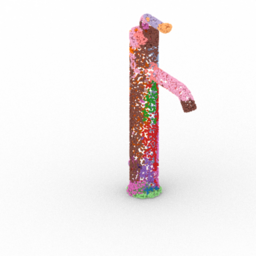}  &
        \includegraphics[{width=.101\linewidth}]{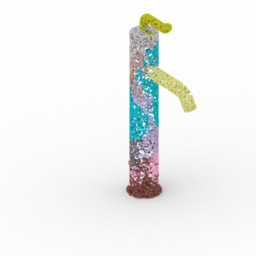}  &
        \includegraphics[{width=.101\linewidth}]{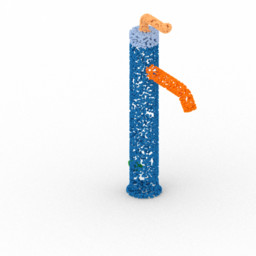}  &
        \includegraphics[{width=.101\linewidth}]{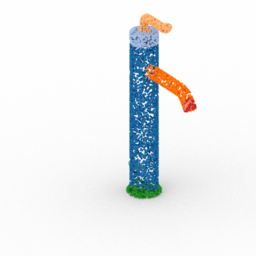}  &
        \includegraphics[{width=.101\linewidth}]{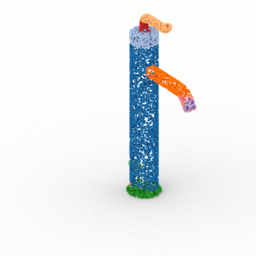}  &
        \includegraphics[{width=.101\linewidth}]{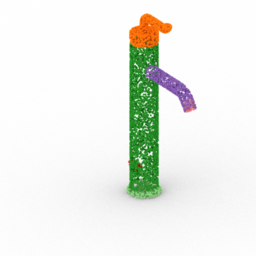} \\
                \includegraphics[trim={0.0cm 2.0cm 0.0cm 0.0cm},clip,{width=.101\linewidth}]{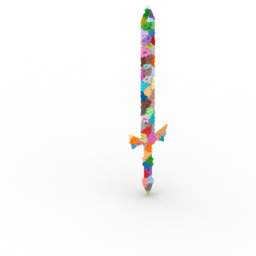}  &
        \includegraphics[trim={0.0cm 1.5cm 0.0cm 0.0cm},clip,{width=.101\linewidth}]{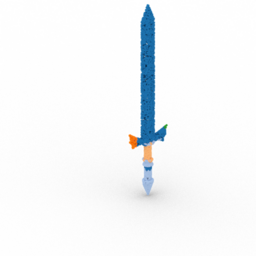}  &
        \includegraphics[trim={0.0cm 1.5cm 0.0cm 0.0cm},clip,{width=.101\linewidth}]{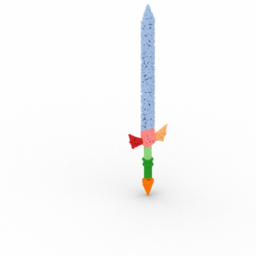}  &
        \includegraphics[trim={0.0cm 1.5cm 0.0cm 0.0cm},clip,{width=.101\linewidth}]{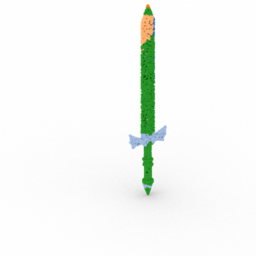}  &
        \includegraphics[trim={0.0cm 1.5cm 0.0cm 0.0cm},clip,{width=.101\linewidth}]{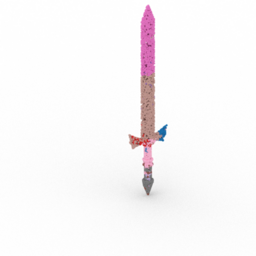}  &
        \includegraphics[trim={0.0cm 1.5cm 0.0cm 0.0cm},clip,{width=.101\linewidth}]{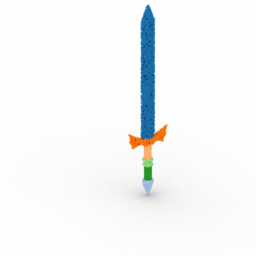}  &
        \includegraphics[trim={0.0cm 1.5cm 0.0cm 0.0cm},clip,{width=.101\linewidth}]{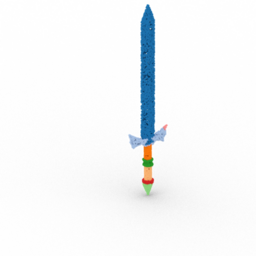}  &
        \includegraphics[trim={0.0cm 1.5cm 0.0cm 0.0cm},clip,{width=.101\linewidth}]{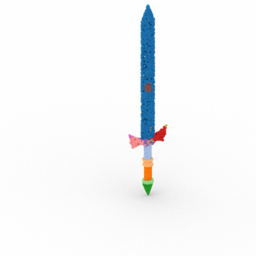}  &
        \includegraphics[trim={0.0cm 1.5cm 0.0cm 0.0cm},clip,{width=.101\linewidth}]{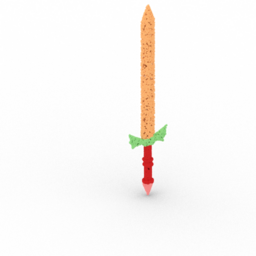}  \\
                \includegraphics[{width=.101\linewidth}]{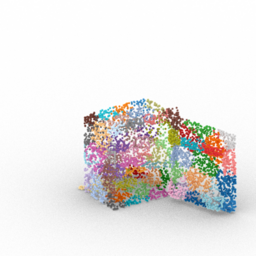}  &
        \includegraphics[{width=.101\linewidth}]{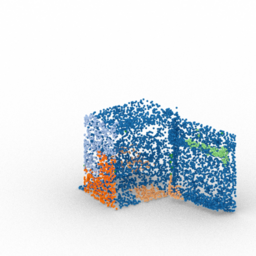}  &
        \includegraphics[{width=.101\linewidth}]{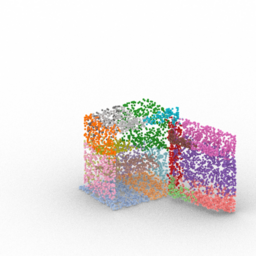}  &
        \includegraphics[{width=.101\linewidth}]{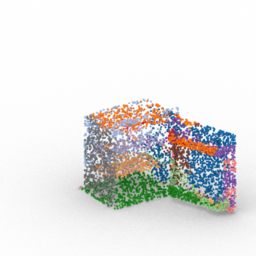}  &
        \includegraphics[{width=.101\linewidth}]{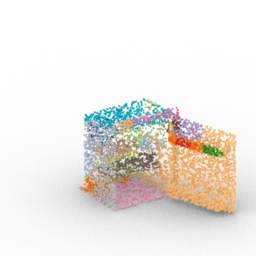}  &
        \includegraphics[{width=.101\linewidth}]{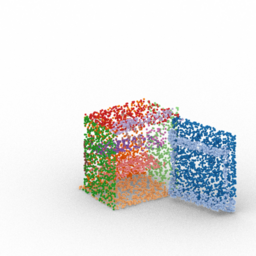}  &
        \includegraphics[{width=.101\linewidth}]{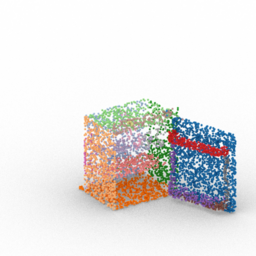}  &
        \includegraphics[{width=.101\linewidth}]{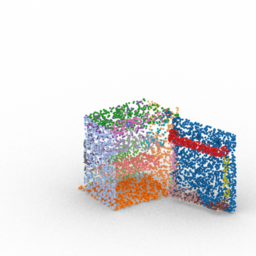}  &
        \includegraphics[{width=.101\linewidth}]{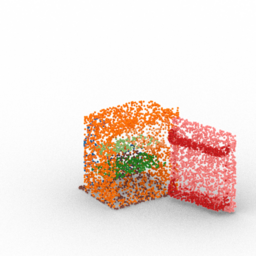} \\
                \includegraphics[{width=.101\linewidth}]{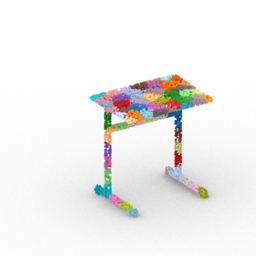}   &
        \includegraphics[{width=.101\linewidth}]{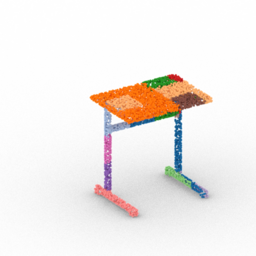}  &
        \includegraphics[{width=.101\linewidth}]{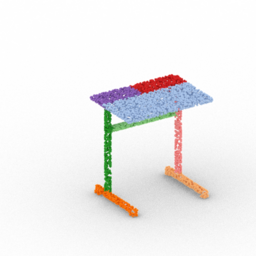}  &
        \includegraphics[{width=.101\linewidth}]{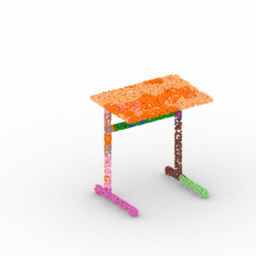}  &
        \includegraphics[{width=.101\linewidth}]{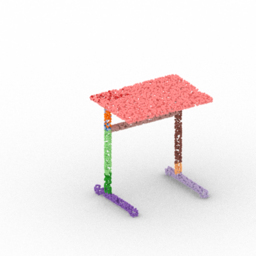}  &
        \includegraphics[{width=.101\linewidth}]{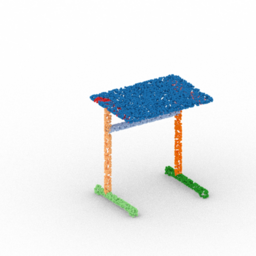}  &
        \includegraphics[{width=.101\linewidth}]{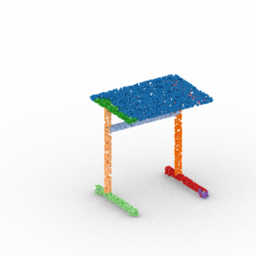}  &
        \includegraphics[{width=.101\linewidth}]{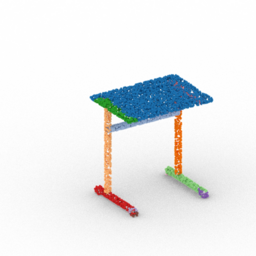}  &
        \includegraphics[{width=.101\linewidth}]{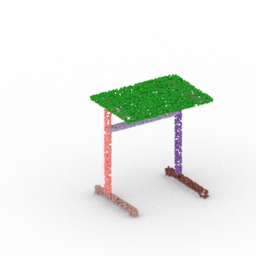} \\

    \end{tabular}
    \caption{ Additional qualitative region decomposition comparisons between SHRED and baseline methods. In columns 6-8, we show how varying SHRED's merge-threshold changes the granularity of the output decomposition.}
    \label{fig:qual_seg_comp}
\end{figure*}

\end{document}